\documentclass[letterpaper, 10pt, conference]{ieeeconf}
\IEEEoverridecommandlockouts

\usepackage{bm}
\usepackage{cite}
\usepackage{amsmath,amssymb,amsfonts}
\usepackage{algorithmic}
\usepackage{graphicx}
\usepackage{textcomp}
\usepackage{xcolor}
\usepackage{comment}
\usepackage{microtype}
\usepackage{subcaption}
\let\labelindent\relax
\usepackage{enumitem}

\begin{document}

\AddToShipoutPictureBG*{%
  \AtPageUpperLeft{%
    \setlength{\unitlength}{1mm}%
    \put(0,-12){\makebox(\paperwidth,0)[c]{\parbox{0.8\textwidth}{\centering\textcolor{gray}{\large This paper has been accepted for publication at the 2025 IEEE-RAS 24th International Conference on Humanoid Robots (Humanoids), Seoul, 2025.  ©IEEE}}}}
  }
}
\title{\bf \LARGE 
\resizebox{\textwidth}{!}{A Framework for Optimal Ankle Design of Humanoid Robots}
\vspace{-1.5em}
}

\author{
Guglielmo Cervettini$^{1,2,\star}$,
Roberto Mauceri$^{1,2,\star}$, 
Alex Coppola$^{1}$,
Fabio Bergonti$^{1}$, \\
Luca Fiorio$^{3}$,
Marco Maggiali$^{3}$,
Daniele Pucci$^{1}$
\vspace{-1.0em}
\thanks{$^\star$ \textbf{The two authors equally contributed to the paper.}}
\thanks{
$^{1}$Artificial and Mechanical Intelligence Laboratory, Istituto Italiano di Tecnologia (IIT).
$^{2}$School of Computer Science, University of Manchester.
$^{3}$iCub Tech Facility, Istituto Italiano di Tecnologia (IIT).
}
}
\maketitle
\begin{abstract}
The design of the humanoid ankle is critical for safe and efficient ground interaction. Key factors such as mechanical compliance and motor mass distribution have driven the adoption of parallel mechanism architectures. However, selecting the optimal configuration depends on both actuator availability and task requirements.
We propose a unified methodology for the design and evaluation of parallel ankle mechanisms. A multi-objective optimization synthesizes the mechanism geometry, the resulting solutions are evaluated using a scalar cost function that aggregates key performance metrics for cross-architecture comparison.
We focus on two representative architectures: the Spherical-Prismatic-Universal (SPU) and the Revolute-Spherical-Universal (RSU). For both, we resolve the kinematics, and for the RSU, introduce a parameterization that ensures workspace feasibility and accelerates optimization.
We validate our approach by redesigning the ankle of an existing humanoid robot. The optimized RSU consistently outperforms both the original serial design and a conventionally engineered RSU, reducing the cost function by up to 41\% and 14\%, respectively.
\vspace{-0.5em}
\end{abstract}

\section{Introduction}
\label{sec:introduction}
Modern humanoid robots are increasingly envisioned as versatile agents capable of operating in human-centric environments and executing complex locomotion and manipulation tasks with a high degree of \emph{agility} and \emph{dexterity}. These capabilities are essential not only for navigating unstructured terrains and ensuring safe task execution in real-world scenarios, but also for achieving levels of dynamism that can rival or exceed human performance. Realizing such functionality requires not only continued advancements in control algorithms but also fundamental improvements in mechanical design, particularly in the limbs that interact directly with the environment~\cite{kim2024hyperleg}.

Two key factors critically influence \emph{agility}: the distribution of mass throughout the robot and the mechanical compliance of its joints. A higher placement of the center of mass (CoM), especially by reducing distal masses in the lower body, improves energy efficiency during movement~\cite{Choi2006}. In this context, parallel mechanisms architectures offer a crucial advantage by allowing heavy actuators to be positioned proximally~\cite{gim2022implementation,zhou2018comprehensive}.
Mechanical compliance — specifically, high backdrivability — enables safer and more effective interaction with the environment. Enhanced compliance improves both robustness to unexpected terrain asperities and the ability to dissipate energy safely during dynamic movements~\cite{alfayad2009new}. These properties are crucial to achieving more forgiving gait patterns and reducing the risk of damage during high-agility tasks.

Among all limb joints, the ankle is especially influential in locomotion, as it is usually the first to respond to ground interactions. For this reason, improving its mechanical design can yield significant gains in agility. Reflecting this importance, recent years have seen a widespread adoption of parallel mechanisms specifically for ankle joints in next-generation humanoid robots. Notable examples include 
Tesla's Optimus, Unitree's G1, PNDbotics' Adam, Agility Robotics' Digit, Fourier's N1, LOLA~\cite{lohmeier2009humanoid}, Pandora~\cite{fuge2023design}, and Kangaroo~\cite{roig2022hardware}. 

Parallel mechanisms offer several inherent advantages over serial architectures: increased structural rigidity, higher positioning precision through averaging of chain errors, reduced actuator size requirements due to torque sharing, and improved dynamics from the relocation of heavy components to more proximal frames~\cite{perera2024staccatoe}.
However, parallel structure design presents significant challenges. It demands careful selection of a larger set of geometric parameters and must account for substantial variations in joint torques and velocities across the workspace~\cite{gim2022implementation}. Unlike serial chains, where performance metrics can often be localized to specific joints, parallel architectures require a global, coupled optimization~\cite{shin2022design}.
\begin{figure}
    \centering \includegraphics[width=0.7\linewidth]{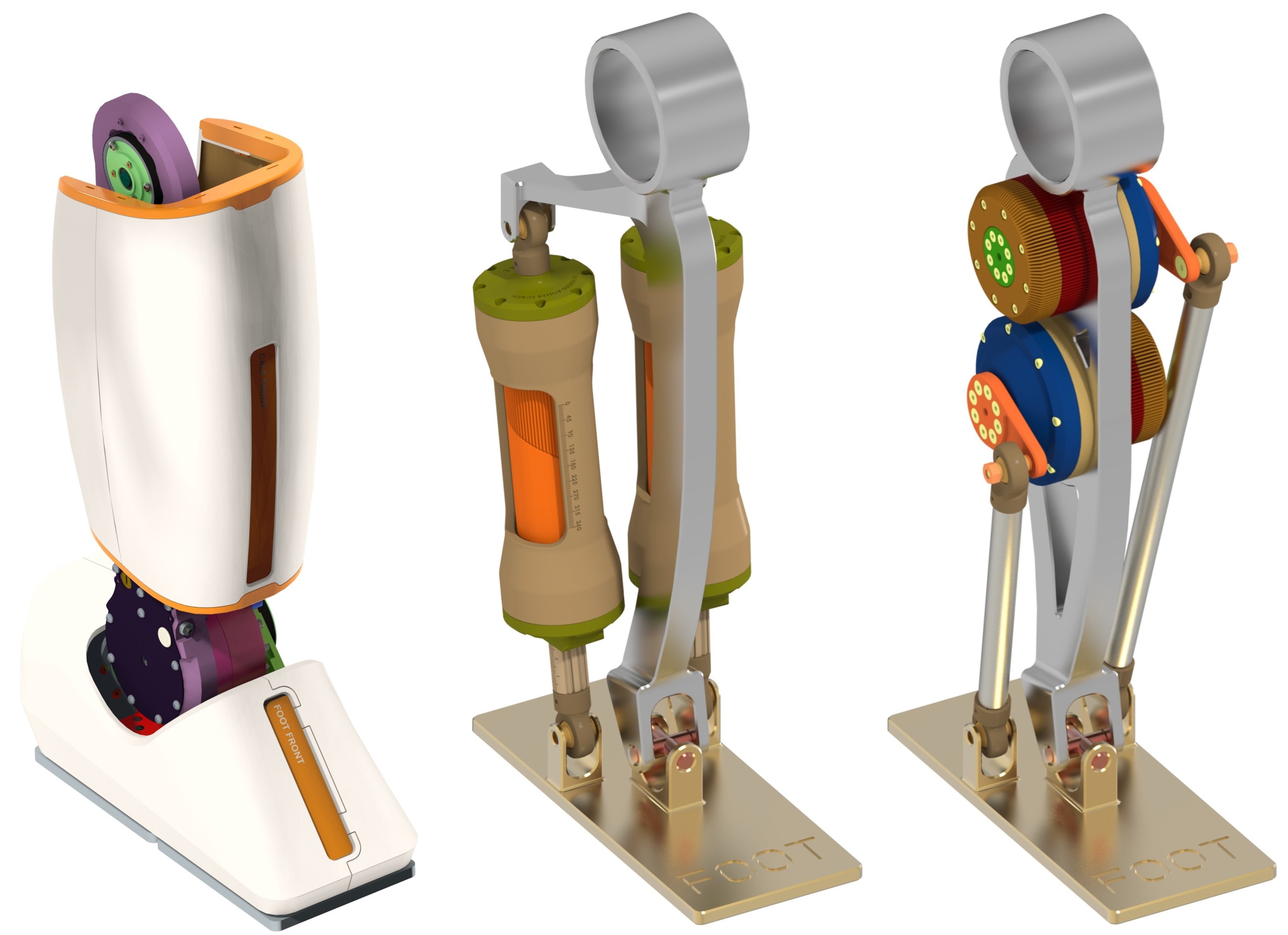}
    \vspace{0.5em}
    \caption{Examples of two-degrees-of-freedom ankle mechanisms. From left to right: serial, SPU and RSU architectures.}
    \label{fig:SPU_RSU_render}
    \vspace{-1.5em}
\end{figure}

Geometric parameter synthesis of parallel mechanisms can be tackled with optimization algorithms~\cite{WU20141377, YANG2022104725, semini2016design, ivolga2023computational, pierrot2009optimal}. However, their application to ankle mechanisms remains limited. Choosing the kinematic structure itself is equally critical. Several surveys outline possible structures~\cite{hashimoto2020mechanics, ficht2021bipedal}, but this choice is highly context-dependent, influenced by target locomotion tasks, available actuators, and design constraints. As a result, these surveys lack quantitative criteria for identifying the most appropriate solution. Consequently, designers currently lack a systematic tool to support both architecture selection and parameter tuning, guided by quantitative performance indicators tailored to specific applications.

Here, we present a unified framework for the design and comparative evaluation of two types of kinematics structures: the Spherical-Prismatic-Universal (SPU) mechanism with linear actuators, and the Revolute-Spherical-Universal (RSU) mechanism with rotary actuators, depicted in Fig.\ref{fig:SPU_RSU_render}. Our methodology identifies the optimal actuator configuration, kinematic structure, and geometric parameters through a multi-objective optimization process, combined with a cost-function-based evaluation strategy that enables cross-architecture and cross-actuator comparisons. The method relies on kinematic modeling and a novel RSU parameterization that ensures workspace feasibility.\looseness=-1

\section{Background}
\label{sec:background}
\subsection{Notation}
\begin{itemize}
    \item Geometric points in the 3D Euclidean space are denoted by uppercase upright letters $\mathrm{P} \in \mathbb{E}^3$, while geometric vectors in the 3D Euclidean vector space are denoted by bold lowercase letters $\bm{v} \in \mathbb{\vec{E}}^3$. The subtraction of two points results in a vector, i.e. $(\mathrm{Q} - \mathrm{P}) \in \vec{\mathbb{E}}^3$.
    \item The Euclidean norm of a vector $\bm{v}$ is denoted by $\| \bm{v} \| \in \mathbb{R}$. The notation $\hat{\bm{v}}$ indicates a unit vector, i.e. $\| \hat{\bm{v}} \| \triangleq 1$.
    \item Frames are designated by uppercase calligraphic letters $\mathcal{A}\triangleq (\hat{\bm{x}}_\mathcal{A},\, \hat{\bm{y}}_\mathcal{A},\, \hat{\bm{z}}_\mathcal{A})$. We only use orthonormal frames.
    \item Coordinates of vectors expressed with respect to (w.r.t.) a frame $\mathcal{A}$ are indicated as ${}^{\mathcal{A}}\bm{v} \triangleq [v_x,\, v_y,\, v_z]^\top\! \in \mathbb{R}^3$.
    \item ${}^{\mathcal{A}}R_{\mathcal{B}} \in \mathrm{SO}(3)$ is the rotation matrix aligning $\mathcal{A}$ with $\mathcal{B}$. It can also be interpreted as the coordinate transformation from $\mathcal{B}$ to $\mathcal{A}$. The symbols $R_x,\, R_y,\, R_z \in \mathrm{SO}(3)$ denote the elementary rotations around the coordinate axes.
\end{itemize}

\subsection{Parallel Mechanisms}
\emph{Parallel mechanisms} represent a distinct class of mechanical systems characterized by their closed-loop architecture, wherein a moving platform, or end-effector (EE), is connected to a fixed base, or world (W), through multiple kinematic chains~\cite{huang2012theory}. \looseness=-1

\subsubsection{Closed Kinematic Chains}
The foundation of parallel mechanisms lies in the concept of \emph{closed kinematic chains}. A closed kinematic chain is defined as any kinematic chain that contains one or more loops in its connectivity graph representation~\cite{yamane2019closed} (see Fig.\ref{fig:topological_graphs}).

Parallel mechanisms represent a specific application of closed kinematic chains. They are constructed as closed chains comprising a fixed base and a moving platform connected by a set of ``legs", that are typically serial kinematic chains (Fig.\ref{fig:topological_graphs}). The \emph{loop closure equations}, which are essential for the kinematic analysis of these mechanisms, directly arise from the presence of these closed chains. In particular, the inverse kinematics (IK) problem, derived from these equations, is generally nonlinear and may admit no solution or multiple solutions \cite{liu2014parallel}. \looseness=-1

\subsubsection{Jacobian Matrix}
The Jacobian matrix is a fundamental tool in the differential analysis of mechanism motion~\cite{lynch2017modern}. Mathematically, it provides a linear relationship between: (i) the joint velocities $\dot{\bm{q}}$ and the resulting twist $\mathbf{v}$ (linear and angular velocities) of the EE $\mathbf{v} \triangleq J(\bm{q})\, \dot{\bm{q}}$, (ii) the wrench $\mathbf{f}$ (forces and torques) applied at the EE and the corresponding forces or torques $\bm{\tau}$ of the actuators $\bm{\tau} = J(\bm{q})^\top\, \mathbf{f}$. The Jacobian matrix explicitly depends on the system configuration $\bm{q}$.

The Jacobian matrix provides a powerful mathematical tool for identifying singular configurations~\cite{gosselin1990singularity}. A key indicator of a singularity is when the rank of the Jacobian matrix drops below its maximum possible value. A singular Jacobian matrix implies that there are certain directions of EE motion that cannot be achieved by any finite joint velocities, and conversely, unbounded actuator forces could be required to achieve a finite wrench at the EE. Kinematic inversion also becomes ill-posed or impossible at singularities. Therefore, understanding and avoiding singular configurations are crucial for the safe and effective operation of parallel mechanisms.

\subsubsection{Manipulability Ellipsoid}
The manipulability ellipsoid is a valuable geometric construct used to assess a mechanism's capability to move or exert force in different directions from a given configuration \cite{yoshikawa1985manipulability}. It provides a visual representation of the mechanism's dexterity and force capabilities. A manipulability ellipsoid that is closer to a sphere in shape suggests a more uniform motion capability in all directions, a property known as isotropy. In contrast, an ellipsoid with a more flattened or elongated shape indicates anisotropic behavior, meaning the mechanism is better suited for motion or force exertion along certain axes than others. The manipulability ellipsoid is mathematically derived from the Jacobian matrix of the mechanism as $M(\bm{q}) \triangleq J(\bm{q})\, J(\bm{q})^\top$. The eigenvalues and eigenvectors of $M(\bm{q})$ define the shape and orientation of the ellipsoid, respectively. The \emph{manipulability ratio}, defined as the ratio between the largest $\lambda_{\text{max}}$ and the smallest $\lambda_{\text{min}}$ eigenvalues of $M(\bm{q})$ serves as a useful indicator of the mechanism’s manipulability \cite{lynch2017modern}: \looseness=-1
\begin{equation}
    \kappa \triangleq \sqrt{\lambda_\text{max}\,/\,\lambda_\text{min}}.
    \label{eq:manipulability_ratio}
\end{equation}

\section{Kinematic Analysis of the Mechanisms}
\label{sec:kinematic_analysis}
In this section, we will analyze the geometry of two parallel mechanisms. First, we will introduce the topological graphs of the mechanisms under study. We will then move on to a geometric representation that will allow us to write the loop closure equations and derive the IK closed-form expression.\looseness=-1 

\begin{figure}[!t]
    \vspace{0.5em}
    \centering   
    \includegraphics[width=0.85\linewidth]{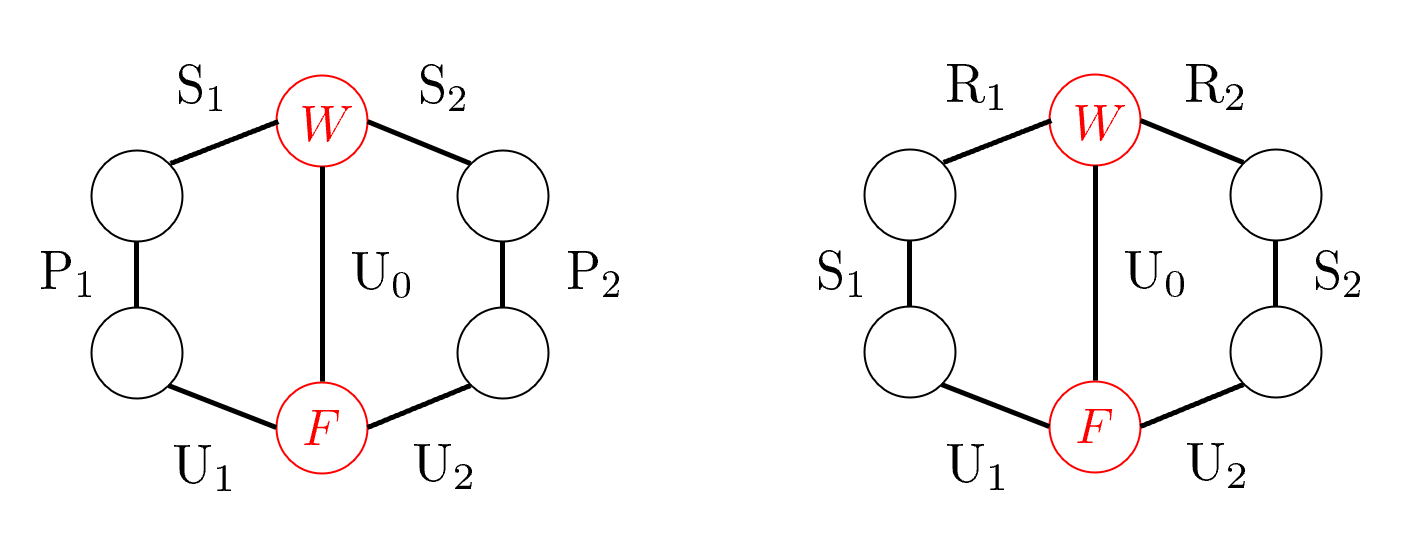}
    \caption{Connectivity graphs of SPU and RSU mechanisms. Links are represented as nodes and joints as edges. Each mechanism has 3 ``legs" connecting the world W to the end-effector F (foot). Each graph contains 2 independent loops.\looseness=-1}
    \label{fig:topological_graphs}
    \vspace{-1.5em}
\end{figure}

In this context, the role of the fixed base for the mechanism will be played by the shin, while the foot will serve as the end-effector. The shin $W$ and the foot $F$ are connected to each other through the interposition of 3 legs (Fig. \ref{fig:topological_graphs}). The central leg consists of a single universal joint $\mathrm{U}_0$ that directly connects the shin and the foot. The angles $(\varphi,\, \vartheta)$ associated to this joint define the orientation of the foot w.r.t. the shin. The other two legs, which are topologically equivalent to each other and indexed by the letter $i \in \{1,\, 2 \}$, are more complex and characterize the mechanisms under consideration. The SPU mechanism has, as its leg $i$, a serial kinematic chain connecting four links through the sequence of joints: spherical $\mathrm{S}_i$, prismatic $\mathrm{P}_i$, and universal $\mathrm{U}_i$. The RSU mechanism has, as its leg $i$, a serial kinematic chain connecting four links through the sequence of joints: revolute $\mathrm{R}_i$, spherical $\mathrm{S}_i$, and universal $\mathrm{U}_i$.\looseness=-1

The degrees of mobility of these two spatial mechanisms are equal to $2$ and, in this case, can be evaluated using the Grübler-Kutzbach criterion~\cite{gogu2005chebychev}. Consequently, the actuation of only two joints is required to fully control the EE pose. \looseness=-1

A natural choice is to actuate the joints with a single degree of freedom. Specifically, in the SPU mechanism, control is achieved via the elongations of the prismatic joints, denoted as $\zeta_1$ and $\zeta_2$, while in the RSU mechanism, it is governed by the angles of the revolute joints, $\alpha_1$ and $\alpha_2$. $\mathcal{W}$ is the world frame and $\mathcal{F}$ is the foot frame. The orientation of $\mathcal{F}$ w.r.t. $\mathcal{W}$ is defined by the roll and pitch angles $(\varphi,\, \vartheta)$ and the corresponding transformation is given by ${}^\mathcal{W}R_\mathcal{F}(\varphi,\, \vartheta) \triangleq R_y(\vartheta) R_x(\varphi)$.
\begin{figure*}[ht]
\vspace{-0.5em}
    \centering
     \begin{subfigure}[b]{0.49\textwidth}
        \centering
        \includegraphics[width=0.78\linewidth]{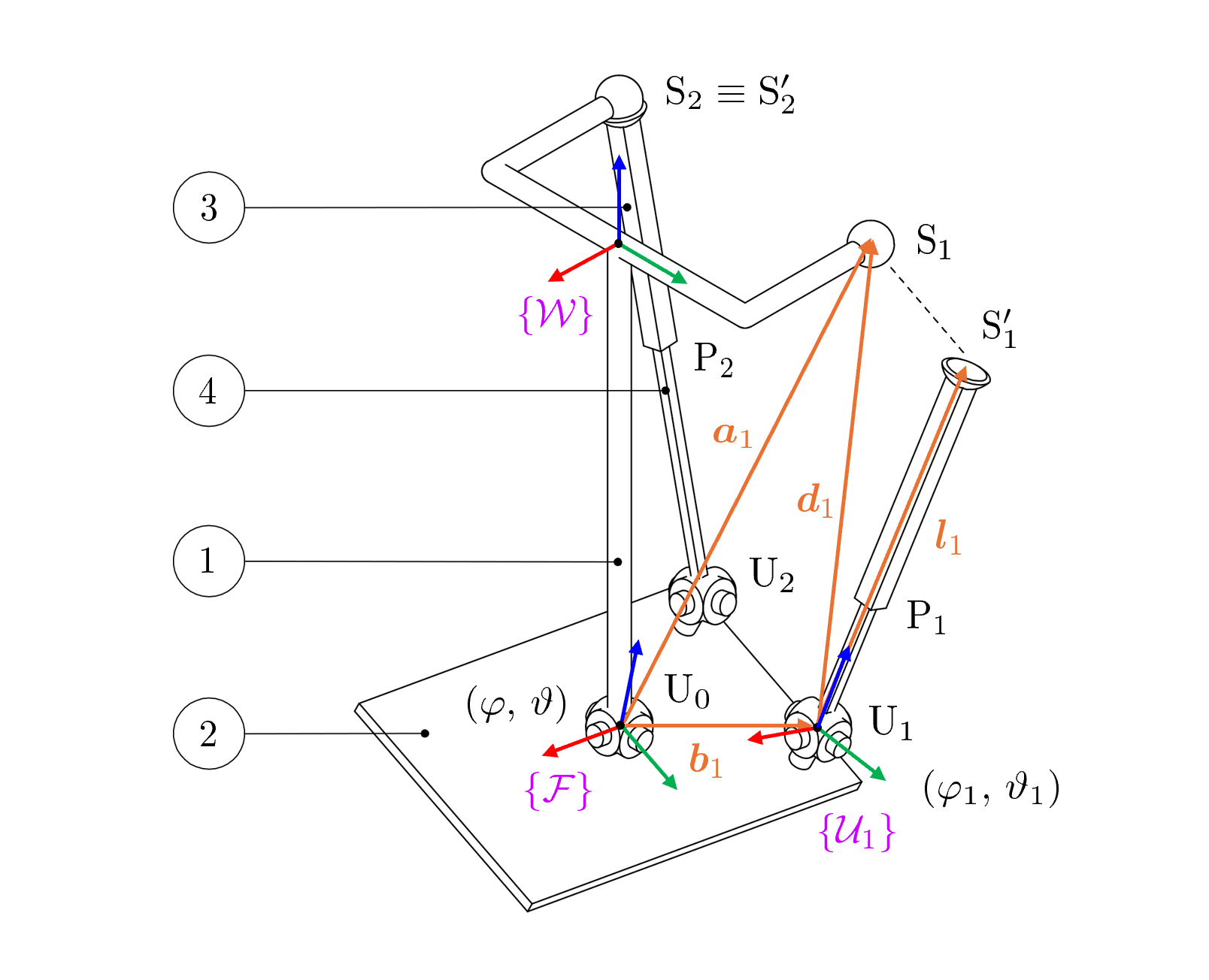}
        \vspace{-5pt}
        \caption{SPU mechanism: 1) shin, 2) foot, 3) slider, 4) linear guide.}
         \label{fig:diagram_SPU}
    \end{subfigure}
    \begin{subfigure}[b]{0.49\textwidth}
    \centering
        \includegraphics[width=0.78\linewidth]{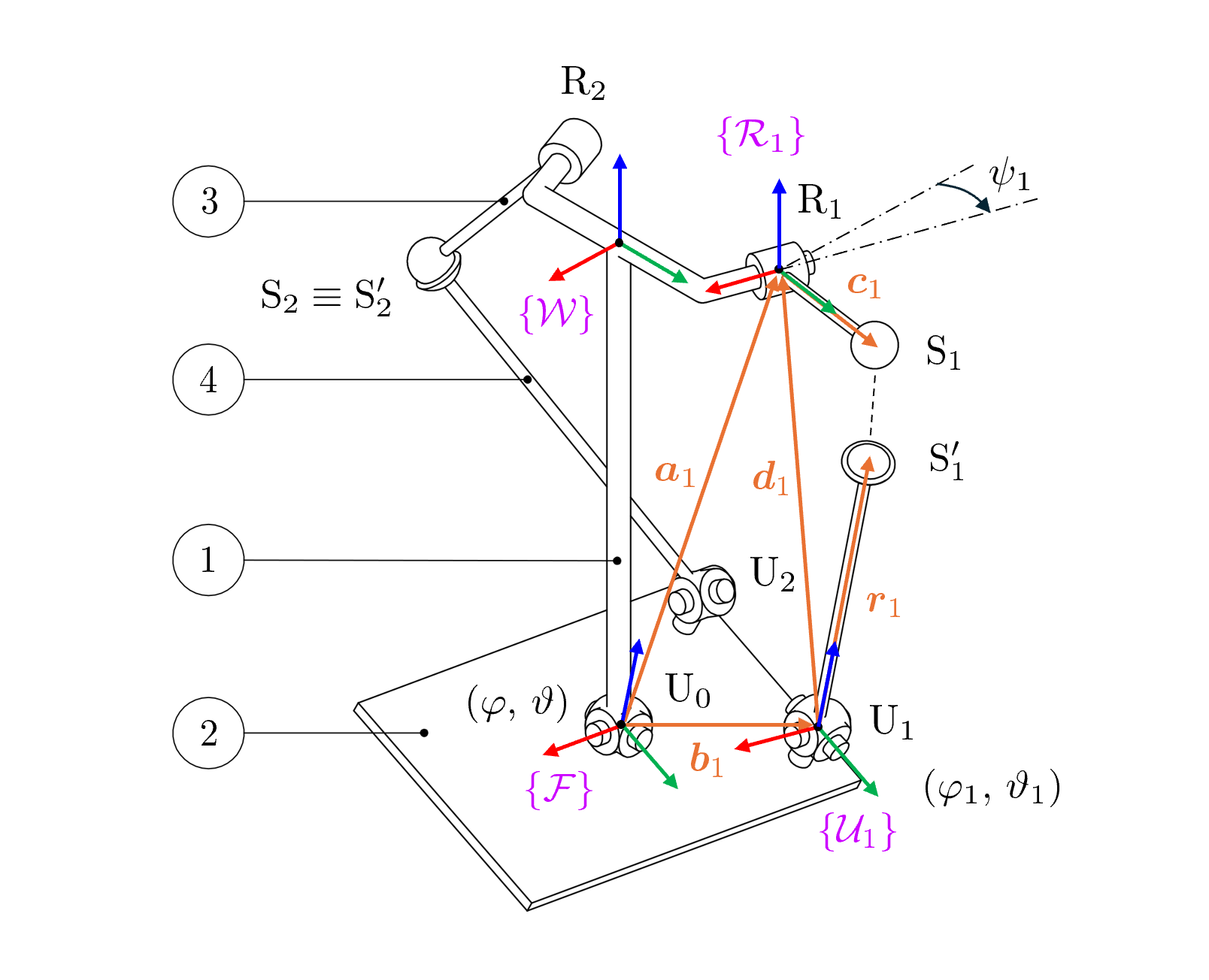}
        \vspace{-5pt}
        \caption{RSU mechanism: 1) shin, 2) foot, 3) crank, 4) rod.}
        \label{fig:diagram_RSU}
    \end{subfigure}
    \caption{Kinematic diagrams of the SPU (a) and RSU (b) mechanisms.}
    \label{fig:diagrams_SPU_RSU}
    \vspace{-1.5em}
\end{figure*}
Both mechanisms feature a spherical joint within their kinematic loops. It is convenient to open the mechanism at the level of this joint, as this approach simplifies the kinematic closure by imposing the coincidence of two points $\mathrm{S}_i$ and $\mathrm{S}'_i$ (3 equations for each independent kinematic loop). 

\subsection{SPU Parallel Mechanism - Closed-form IK}
\label{subsec:ikSPU}
Considering the use of actuated prismatic joints, the IK problem is formulated as the determination of the joint linear displacements $\bm{q} = [\zeta_1,\, \zeta_2]^\top\!$ required to achieve a desired foot orientation w.r.t. $\mathcal{W}$ in terms of roll and pitch angles $\varphi,\, \vartheta$. In the SPU mechanism, depicted in Fig.\ref{fig:diagram_SPU}, we define: 
\begin{equation*}
\begin{array}{lcl}
    \bm{a}_i \triangleq \mathrm{S}_i - \mathrm{U}_0, & &
    \bm{d}_i \triangleq \mathrm{S}_i - \mathrm{U}_i, \\
    \bm{b}_i \triangleq \mathrm{U}_i - \mathrm{U}_0, & &
    \:\bm{l}_i \triangleq \mathrm{S}'_i - \mathrm{U}_i.
\end{array}
\end{equation*}
${}^\mathcal{W}\bm{a}_i,\, {}^\mathcal{F}\bm{b}_i \in \mathbb{R}^3$ are constant tuples and represent the geometric parameters of the mechanism we consider in our design. The coordinates of $\bm{d}_i$ and $\bm{l}_i$ w.r.t. $\mathcal{W}$ are:
\begin{equation}
    {}^\mathcal{W}\bm{d}_i(\varphi,\, \vartheta) = {}^\mathcal{W}\bm{a}_i - {}^\mathcal{W}\!R_\mathcal{F}(\varphi,\, \vartheta)\, {}^\mathcal{F}\bm{b}_i,
    \label{eq:d_i_SPU}
\end{equation}
\begin{equation*}
{}^\mathcal{W}\bm{l}_i(\zeta_i,\,\varphi_i,\, \vartheta_i,\, \varphi,\, \vartheta) = {}^\mathcal{W}\!R_\mathcal{F}(\varphi,\, \vartheta)\, {}^\mathcal{F}\!R_{\,\mathcal{U}_i}(\varphi_i,\, \vartheta_i)\, {}^{\mathcal{U}_i}\bm{l}_i(\zeta_i), 
\end{equation*}
with ${}^\mathcal{F}\!R_{\mathcal{U}_i}(\varphi_i,\vartheta_i) {\triangleq} [R_y(\vartheta_i)R_x(\varphi_i)]^\top \!\!$ and ${}^{\mathcal{U}_i}\bm{l}_i(\zeta_i) \triangleq [0,0,\zeta_i]^\top \!\!.$ The angles $\varphi_i$, $\vartheta_i$ are the roll and pitch angles associated to the universal joint $\mathrm{U}_i$.
The loop closure equation is directly derived by imposing $\mathrm{S}'_i \equiv \mathrm{S}_i$, i.e. ${}^\mathcal{W}(\mathrm{S}'_i {-} \mathrm{U}_i) ={}^\mathcal{W}(\mathrm{S}_i {-} \mathrm{U}_i)$:\looseness=-1
\begin{equation}
    {}^\mathcal{W}\bm{l}_i(\zeta_i,\,\varphi_i,\, \vartheta_i,\, \varphi,\, \vartheta) = {}^\mathcal{W}\bm{d}_i(\varphi,\, \vartheta).
    \label{eq:IK_SPU}
\end{equation}
To make the equation independent from the unknown variables $\varphi_i$ and $\vartheta_i$, the Euclidean squared norm is taken on both sides of \eqref{eq:IK_SPU}. Since Euclidean norms are invariant under orthogonal transformations, the reference frame superscript is omitted:
\begin{equation}
    \zeta_i^2 = \| \bm{d}_i (\varphi,\, \vartheta) \|^2,
\end{equation}
with $\| \bm{d}_i(\varphi,\, \vartheta) \|^2 = \| \bm{a}_i \|^2 + \| \bm{b}_i \|^2 - 2\, {}^\mathcal{W}\bm{a}_i^\top\, {}^\mathcal{W}\!R_\mathcal{F}(\varphi,\, \vartheta)\, {}^\mathcal{F}\bm{b}_i$ (derived from \eqref{eq:d_i_SPU}). Finally, the IK results in:
\begin{equation*}
    \zeta_i = \pm \sqrt{\| \bm{a}_i \|^2 + \| \bm{b}_i \|^2 - 2\, {}^\mathcal{W}\bm{a}_i^\top\, {}^\mathcal{W}\!R_\mathcal{F}(\varphi,\, \vartheta)\, {}^\mathcal{F}\bm{b}_i}.
\end{equation*}
There are thus two possible solutions. For design reasons, we adopt the solution with the positive sign. The solution with the negative sign, although mathematically admissible, would correspond to an elongation occurring in the direction opposite to $\hat{\bm{z}}_{\mathcal{U}_i}$, which would intersect with the foot.

\subsection{RSU Parallel Mechanism - Closed-form IK}
\label{subsec:ikRSU}
Considering the use of actuated revolute joints, the IK problem is formulated as the determination of the joint angular displacements $\bm{q} = [\alpha_1,\, \alpha_2]^\top\!$ required to achieve a desired foot orientation w.r.t. $\mathcal{W}$ in terms of roll and pitch angles $\varphi,\, \vartheta$. In the RSU mechanism, depicted in Fig. \ref{fig:diagram_RSU}, we define:
\begin{equation*}
\begin{array}{lclcl}
    \bm{a}_i \triangleq \mathrm{R}_i - \mathrm{U}_0, & &
    \bm{c}_i \triangleq \mathrm{S}_i - \mathrm{R}_i, & & 
    \bm{d}_i \triangleq \mathrm{R}_i - \mathrm{U}_i, \\
    \bm{b}_i \triangleq \mathrm{U}_i - \mathrm{U}_0, & &
    \bm{r}_i \triangleq \mathrm{S}'_i - \mathrm{U}_i. & &
\end{array}
\end{equation*}
${}^\mathcal{W}\bm{a}_i,\, {}^\mathcal{F}\bm{b}_i \in \mathbb{R}^3$ are constant $3$-tuples and represent some of the geometric parameters of the mechanism. The coordinates of $\bm{d}_i$, $\bm{c}_i$ (crank), and $\bm{r}_i$ (rod) w.r.t. $\mathcal{W}$ are:
\begin{equation}
    {}^\mathcal{W}\bm{d}_i(\varphi,\, \vartheta) = {}^\mathcal{W}\bm{a}_i - {}^\mathcal{W}\!R_\mathcal{F}(\varphi,\, \vartheta)\, {}^\mathcal{F}\bm{b}_i,
    \label{eq:d_i_RSU}
\end{equation}
\begin{equation*}
    {}^\mathcal{W}\bm{c}_i(\alpha_i) = {}^\mathcal{W}\!R_{\mathcal{R}_i}(\alpha_i)\, {}^{\mathcal{R}_i}\bm{c}_i,
\end{equation*}
\begin{equation*}
    {}^\mathcal{W}\bm{r}_i(\varphi_i,\, \vartheta_i,\, \varphi,\, \vartheta) = {}^\mathcal{W}\!R_\mathcal{F}(\varphi,\, \vartheta)\, {}^\mathcal{F}\!R_{\,\mathcal{U}_i}(\varphi_i,\, \vartheta_i)\, {}^{\mathcal{U}_i}\bm{r}_i,
\end{equation*}
where ${}^\mathcal{F}\!R_{\,\mathcal{U}_i}(\varphi_i,\, \vartheta_i) = [R_y(\vartheta_i)\, R_x(\varphi_i)]^\top$, ${}^{\mathcal{U}_i}\bm{r}_i = [0,\, 0,\, r_i]^\top$, ${}^\mathcal{W}\!R_{\mathcal{R}_i}(\alpha_i) = R_z(\psi_i)\, R_x(\alpha_i)$, and ${}^{\mathcal{R}_i}\bm{c}_i = [0,\, c_i,\, 0]^\top$. The angles $\varphi_i$, $\vartheta_i$ are the roll and pitch angles associated to the universal joint $\mathrm{U}_i$. The constant quantities $c_i,\, r_i \in \mathbb{R}^+$, and $\psi_i \in \mathbb{S}^1$ are the remaining geometric parameters of the mechanism we consider in our design. The quantities $c_i$ and $r_i$ represent, respectively, the crank and rod lengths, while the angle $\psi_i$ is the rotation of the rotary actuator axis about the vertical axis of the world frame.

The loop closure equation is directly derived by imposing $\mathrm{S}'_i \equiv \mathrm{S}_i$, i.e. ${}^\mathcal{W}(\mathrm{S}'_i - \mathrm{U}_i) = {}^\mathcal{W}(\mathrm{S}_i - \mathrm{R}_i) + {}^\mathcal{W}(\mathrm{R}_i - \mathrm{U}_i)$:
\begin{equation}
{}^\mathcal{W}\bm{r}_i(\varphi_i,\, \vartheta_i,\, \varphi,\, \vartheta) = {}^\mathcal{W}\bm{c}_i(\alpha_i) + {}^\mathcal{W}\bm{d}_i(\varphi,\, \vartheta).
\label{eq:IK_RSU}
\end{equation}
To make the equation independent from the unknown variables $\varphi_i$ and $\vartheta_i$, the Euclidean squared norm is taken on both sides of \eqref{eq:IK_RSU}. Since Euclidean norms and dot products are invariant under orthogonal transformations, the reference frame superscript is omitted:\looseness=-1
\begin{equation}
    r_i^2 = c_i^2 + \| \bm{d}_i(\varphi,\, \vartheta) \|^2 + 2\, \bm{d}_i(\varphi,\, \vartheta)^\top \bm{c}_i(\alpha_i),
    \label{eq:IK_RSU_scalar}
\end{equation}
with $\| \bm{d}_i(\varphi,\, \vartheta) \|^2 = \| \bm{a}_i \|^2 + \| \bm{b}_i \|^2 - 2\, {}^\mathcal{W}\bm{a}_i^\top\, {}^\mathcal{W}\!R_\mathcal{F}(\varphi,\, \vartheta)\, {}^\mathcal{F}\bm{b}_i$ (derived from \eqref{eq:d_i_RSU}).
Building on \eqref{eq:IK_RSU_scalar}, the corresponding expression is derived:\looseness=-1
\begin{equation}
     \hat{\bm{d}}_i^{\,\top} \hat{\bm{c}}_i(\alpha_i) = (r_i^2 - c_i^2 - \| \bm{d}_i \|^2)\,/\,(2\, c_i\, \| \bm{d}_i \|) \triangleq k_i.
     \label{eq:k_i}
\end{equation}
To proceed with calculation, world coordinates are used:
\begin{equation*}
    {}^\mathcal{W}\hat{\bm{d}}_i^{\,\top}\, R_z(\psi_i)\, R_x(\alpha_i)\, [0,\, 1,\, 0]^\top = k_i.
\end{equation*}
Defining $\tilde{\bm{d}}_i^{\,\top} \triangleq [\tilde{d}_{x,\, i},\, \tilde{d}_{y,\, i},\, \tilde{d}_{z,\, i}] \triangleq {}^\mathcal{W}\hat{\bm{d}}_i^{\,\top}\, R_z(\psi_i)$, we obtain:
\begin{equation*}
    \tilde{d}_{y,\, i}\, \cos{\alpha_i} + \tilde{d}_{z,\, i}\, \sin{\alpha_i} = k_i.
\end{equation*}
A linear combination of sine and cosine, as shown above, can be represented in polar form as follows:
\begin{equation*}
    \rho_i\, \sin (\alpha_i + \phi_i) = k_i,
\end{equation*}
where $\rho_i \triangleq \sqrt{\tilde{d}_{y,\, i}^{\,2} + \tilde{d}_{z,\, i}^{\,2}}$ and $\phi_i \triangleq \text{arctan2} ( \tilde{d}_{y,\, i},\, \tilde{d}_{z,\, i})$.
Therefore, the two possible solutions of the IK are:
\begin{equation*}
    \alpha_i = -\phi_i + \arcsin \left ( \tfrac{k_i}{\rho_i} \right )
    \; \text{or} \;
    \alpha_i = -\phi_i + \pi - \arcsin \left ( \tfrac{k_i}{\rho_i} \right).
\end{equation*}
The equation, and consequently the IK problem, admits solutions if the following existence condition is satisfied:
\begin{equation*}
    \left | k_i\,/\, \rho_i \right | \leq 1.
\end{equation*} 
If the condition is satisfied as an equality, the two solutions coincide. When \( k_i / \rho_i = 1 \), the unique solution is \(\pi/2 - \phi_i\). When \( k_i / \rho_i = -1 \), the unique solution is \(-\pi/2 - \phi_i\). In both cases, this corresponds to the configuration where the crank and the rod are perfectly aligned, which can result in either an overlapping or non-overlapping condition for the two links. When it is satisfied with an inequality, there are two distinct solutions (positioned symmetrically w.r.t. the axis inclined as $\pi/2 -\phi_i$). If the condition is not satisfied, no feasible solution exists. In other words, due to the given geometric parameters and the desired foot orientation, it is not possible to close the mechanism by connecting the crank and the rod through the spherical joint. The existence condition can be also reformulated by using \eqref{eq:k_i} as:
\begin{equation}
    \left|r_i^2 - c_i^2 - \| \bm{d}_i \|^2\right| \leq 2\, c_i\, \| \bm{d}_i \|\, \rho_i.
    \label{eq:existence_condition}
\end{equation}

\subsection{Reparameterization of the RSU Mechanism}
\label{subsec:RSUparamNew}
When the geometric parameters of the mechanism are chosen arbitrarily, there is no guarantee that the existence condition \eqref{eq:existence_condition} is satisfied for all desired values of foot pitch and roll. It would therefore be beneficial to perform a reparameterization of the mechanism such that this condition is inherently fulfilled. In other words, the desired operational region for the foot should be entirely contained within the region where the IK problem admits solutions (configuration space of the mechanism), regardless of the specific values assigned to the new parameters (Fig. \ref{fig:reparameterization}).\looseness=-1

\subsubsection{Derive the lower and upper bounds for the crank length}
We want to rewrite \eqref{eq:existence_condition} in such a way we can make the crank length $c_i$ explicit. First of all, it is possible to write:
\begin{equation*}
    c_i \geq | r_i^2 - c_i^2 - \| \bm{d}_i \|^2 | \,/\, (2\, \| \bm{d}_i \|\, \rho_i),
    \quad
    \forall (\varphi,\, \vartheta) \in \Omega,
\end{equation*}
where $\Omega$ is the desired operational region for the mechanism.
Then, we note that  $r_i^2 - c_i^2 = (r_i + c_i)(r_i - c_i)$. The quantity $r_i + c_i$ represents the maximum distance $d_{i,\, \text{max}}$ between the universal joint $\mathrm{U}_i$ and the revolute joint $\mathrm{R}_i$ that can be reached by the closed mechanism, while $r_i - c_i$ represents the minimum distance $d_{i,\, \text{min}}$. These cases occur when the crank and the rod are aligned. It follows that:
\begin{equation*}
    c_i \geq | d_{i,\, \text{max}}\, d_{i,\, \text{min}} - \| \bm{d}_i \|^2 |\, / \, (2\, \| \bm{d}_i \|\, \rho_i),
    \quad
    \forall (\varphi,\, \vartheta) \in \Omega.
\end{equation*}
We can also define the minimum and maximum $\mathrm{U}_i$-$\mathrm{R}_i$ distance that can be reached by the mechanism given a desired operational region $\Omega$ (in general, different from $d_{i,\, \text{min}}$ and $d_{i,\, \text{max}}$):\looseness=-1
\begin{equation*}
\begin{aligned}
    d^\star_{i,\, \text{min}} &\triangleq \underset{(\varphi,\, \vartheta)\, \in\, \Omega} \min \| \bm{d}_i \|  \geq d_{i,\, \text{min}}, \\
    d^\star_{i,\, \text{max}} &\triangleq \underset{(\varphi,\, \vartheta)\, \in\, \Omega} \max \| \bm{d}_i \| \leq d_{i,\, \text{max}}.
\end{aligned}
\end{equation*}
We want to compute the minimum possible value for the crank length $c_{i,\, \text{min}}$, so we impose that $d^\star_{i,\, \text{min}} = d_{i,\, \text{min}}$ and $d^\star_{i,\, \text{max}} = d_{i,\, \text{max}}$ (worst case scenario). At this point, the inequality can be rewritten as: 
\begin{equation*}
    c_i \geq | d^\star_{i,\, \text{max}}\, d^\star_{i,\, \text{min}} - \| \bm{d}_i \|^2 |\, / \, (2\, \| \bm{d}_i \|\, \rho_i),
    \quad
    \forall (\varphi,\, \vartheta) \in \Omega.
\end{equation*}
The parameter \( c_i \) must be greater than all values assumed by the preceding expression as the pitch and roll angles vary. Consequently, it must be greater than or equal to a minimum admissible value \( c_{i,\, \text{min}} \), defined as the maximum value attained by the expression over the considered domain:
\begin{equation*}
    c_i \geq c_{i,\, \text{min}} \triangleq \underset{(\varphi,\, \vartheta)\, \in\, \Omega} \max \frac{|d^\star_{i,\, \text{max}}\, d^\star_{i,\, \text{min}} - \| \bm{d}_i \|^2|}{2\, \| \bm{d}_i \|\, \rho_i}.
\end{equation*}
The inequality does not imply any upper bound for the crank length.
We can introduce a new parameter, $\gamma_i {\in} [0,1]$. When this parameter is equal to 0, the crank length takes on its minimum value; as it approaches 1, the crank length approaches infinity:\looseness=-1
\begin{equation*}
    c_i(\gamma_i) \triangleq \frac{1}{1 - \gamma_i}\, c_{i,\, \text{min}}.
\end{equation*}
\subsubsection{Derive the lower and upper bounds for the rod length}
Now $c_i = c_i(\gamma_i)$ is a known parameter. We want to rewrite \eqref{eq:existence_condition} in such a way we can make the rod length $r_i$ explicit. We can start again from the inequality constraint:
\begin{equation*}
|r_i^2 - c_i^2 - \| \bm{d}_i \|^2|  \leq 2\, c_i\, \| \bm{d}_i \|\, \rho_i, \quad \forall (\varphi,\, \vartheta) \in \Omega. \\
\end{equation*}
By the definition of the absolute value function, the previous inequality is equivalent to the following system:
\begin{equation*}
\begin{matrix}
\begin{cases}
r_i^2 \geq c_i^2 + \| \bm{d}_i \|^2 - 2\, c_i\, \| \bm{d}_i \|\, \rho_i, \\
r_i^2 \leq c_i^2 + \| \bm{d}_i \|^2 + 2\, c_i\, \| \bm{d}_i \|\, \rho_i,
\end{cases} & \forall (\varphi,\, \vartheta) \in \Omega.
\end{matrix}
\end{equation*}
The quantity $r_i^2$ must lie within intervals whose bounds vary as a function of the roll and pitch angles. In order to simultaneously satisfy this infinite set of conditions, it is sufficient for $r_i^2$ to be contained within the intersection of all such intervals. This intersection has the following bounds:
\begin{equation*}
\begin{cases}
    r_i^2 \geq r^2_{i,\, \text{min}} \triangleq \underset{(\varphi,\, \vartheta)\, \in\, \Omega} \max \left\{c_i^2 + \| \bm{d}_i \|^2 - 2\, c_i\, \| \bm{d}_i \|\, \rho_i \right\}, \\
    r_i^2 \leq r^2_{i,\, \text{max}} \triangleq \underset{(\varphi,\, \vartheta)\, \in\, \Omega} \min \left\{c_i^2 + \| \bm{d}_i \|^2 + 2\, c_i\, \| \bm{d}_i \|\, \rho_i \right\}.
\end{cases}
\end{equation*}
Now, $r_i$ can be expressed as a convex combination of $r_{i,\, \text{min}}$ and $r_{i,\, \text{max}}$, using a new parameter $\delta_i \in [0,\, 1]$.  When this parameter is equal to 0, the rod length takes on its minimum value, when it is equal to 1, the rod length takes on its maximum value:
\begin{equation*}
    r_i(\delta_i) = (1 - \delta_i)\, r_{i,\, \text{min}} + \delta_i\, r_{i,\, \text{max}}.  
\end{equation*}

\section{Mechanisms Evaluation Framework}
\label{sec:optimization}
This section outlines the methodology used to enable consistent, cross-architecture comparison of ankle mechanisms equipped with different actuators. The evaluation process comprises two main stages.

In the first stage, described in Section~\ref{subsec:geom_opt}, a dedicated multi-objective optimization is performed for each actuator-architecture pair. The goal is to explore the design space and identify a population of geometrically feasible mechanisms that minimize key performance objectives. This optimization is essential because the relationship between geometric parameters and performance is highly nonlinear and architecture-dependent. Moreover, each actuator imposes unique physical and dynamic constraints — such as size, mounting requirements, and force/velocity limits — which must be reflected in a custom design space (feasible set $\Pi$) for each combination. As a result, the optimization must be repeated independently for every actuator-architecture pair.

In the second stage, detailed in Section~\ref{subsec:cost_function}, we extract high-level performance metrics from the optimized populations. These metrics are formulated to be architecture-agnostic, though their computation depends on the specific kinematics of each mechanism. All metrics are normalized across the full set of candidates to ensure comparability, and aggregated into a scalar cost value. This enables a unified ranking of solutions across all actuator-architecture combinations.

This separation is both computationally and conceptually motivated: the first stage (multi-objective optimization) is the most computationally expensive, while the scalar cost function in the second stage is fast to compute and can be freely re-weighted without repeating the optimization. This decoupling also separates internal actuator effort (forces/velocities) from external task-level performance, enabling application-specific ranking of the same optimized populations.

\subsection{Geometrical Parameters Optimization}
\label{subsec:geom_opt}
The proposed strategy is formally framed as a multi-objective optimization problem, which seeks to identify the elements $\bm{\pi}$ within a feasible set $\Pi$ that achieve the best possible compromise in minimizing a set of objective functions: \looseness=-1
\begin{equation} \label{eq:optimisation-problem}
    \underset{\bm{\pi}\, \in\, \Pi}{\min} \left(f_1(\bm{\pi}),\,  f_2(\bm{\pi}) \right).
\end{equation}
The set of all Pareto optimal solutions, denoted \( \Pi^\star \), defines the \emph{Pareto front}. It represents the set of trade-off solutions among conflicting objectives, where any improvement results in the deterioration of at least one objective.

In our case, the elements $\bm{\pi}$ of the search space consist of vectors gathering the geometric parameters of the mechanism, which uniquely identify its morphology:
\begin{equation*}
\begin{aligned}
    \bm{\pi}_\text{SPU} &\triangleq 
    \begin{bmatrix}
        {}^\mathcal{W}\bm{a}_i^\top, & 
        {}^\mathcal{F}\bm{b}_i^\top
    \end{bmatrix}^\top\!, \\
    \bm{\pi}_\text{RSU} &\triangleq 
    \begin{bmatrix}
        {}^\mathcal{W}\bm{a}_i^\top, & 
        {}^\mathcal{F}\bm{b}_i^\top, & 
        \psi_i, & 
        \gamma_i, & 
        \delta_i
    \end{bmatrix}^\top\!,
\end{aligned}
\end{equation*}
where $\gamma_i$ and $\delta_i$ are the parameterization variables introduced in Sec.~\ref{subsec:RSUparamNew}.
The feasible set $\Pi$ is manually defined by constraints on peak actuators ratings, hardware and physical limits, and collision avoidance. $\Pi$ also takes into account the desired \emph{operational region} $\Omega$, defined as the domain of foot orientations that the mechanism must cover.\looseness=-1

The objectives of our optimization are to minimize both the peak actuator force $f_1(\bm{\pi})$, and the peak actuator velocity $f_2(\bm{\pi})$, evaluated over a set of reference trajectory tasks: \looseness=-1
\begin{equation*}
    \begin{aligned}
        f_1(\bm{\pi}) &\triangleq \underset{k\, \in\, K} \max \left \{ \underset{t\, \in\, T_k} \max \; J(\bm{q}_k(t),\, \bm{\pi})^\top \, \bm{\mathrm{f}}_k(t) 
        \right \}, \\
        f_2(\bm{\pi}) &\triangleq
        \underset{k\, \in\, K} \max \left \{ \underset{t\, \in\, T_k} \max \; J(\bm{q}_k(t),\, \bm{\pi})^{-1} \,\bm{\mathrm{v}}_k(t) \right \},
    \end{aligned}
\end{equation*}
where $K$ denotes the set containing the indices of the reference tasks, and $T_k$ represents the time horizon over which the trajectory $k$ is defined. Reference trajectory tasks are expressed as time series of joint position $\bm{q}(t)$, velocity $\dot{\bm{q}}(t)$, and torque $\bm{\tau}(t)$, derived from simulation or experimental data.

Each element $\bm{\pi}$ is evaluated by computing the actuator performances based on closed-form IK (Sec. \ref{sec:kinematic_analysis}) and Jacobian-based kinematic analysis for each reference task. Performance metrics such as peak force/torque and velocity demands are extracted from this analysis. Feasibility across the entire operational region is verified through geometric constraints: for SPU configurations, actuator stroke limits are enforced, while RSU configurations are parameterized to ensure the existence of a valid inverse kinematic solution.

The multi-objective optimization outputs a Pareto front $\Pi^\star$, which necessitates a subsequent ranking based on higher-level criteria tailored to the final application.

\subsection{Scalar Cost Function for Comparative Evaluation}
\label{subsec:cost_function}

To identify the most promising ankle designs across both SPU and RSU architectures, we introduce a cost function that enables unified post-ranking of the Pareto results. This function aggregates multiple performance metrics into a single scalar cost $\xi$, facilitating direct comparison and final selection.

We shift focus from the internal actuated joints of the mechanism (prismatic or revolute) to the external performance at the ankle joints (roll and pitch), which are directly relevant to the robot's motion tasks. In this context, a set of high-level performance metrics is introduced to evaluate each candidate ankle based on its ability to (i) transmit speed and torque effectively to the ankle joints, (ii) exhibit favorable backdrivability and manipulability characteristics, and (iii) maintain a mechanically compact, lightweight, and high-CoM configuration. \looseness=-1

\subsubsection{Performance Metrics}

Each ankle configuration is evaluated according to seven key metrics:
\begin{enumerate}[label=\roman*.]
    \item \emph{Speed} [rad/s]: roll and pitch angular velocities at the ankle universal joint $\mathrm{U}_0$ resulting from the actuators operating at their nominal linear/angular speed.
    \item \emph{Torque} [Nm]: roll and pitch torques available at the ankle joint $\mathrm{U}_0$ when actuators operate at nominal torque/force.
    \item \emph{Backdriving Torque} [Nm]: roll and pitch torques required at the ankle joint $\mathrm{U}_0$ to overcome static friction at the actuators (using manufacturer-specified data).
    \item \emph{Manipulability Ratio} [--]: as defined in \eqref{eq:manipulability_ratio}.
    \item \emph{Compactness} [mm]: minimum radius of a vertical cylinder enclosing all mechanism points in the neutral configuration ($\varphi = \vartheta = 0$).
    \item \emph{Actuation Mass} [kg]: total mass of the actuator pair (RSU includes estimated crank+rod mass).
    \item \emph{CoM Height} [mm]: vertical distance from ground to the actuators' CoM, computed for $\varphi = \vartheta = 0$ using CAD models, accounting for additional linkage mass in RSU. \looseness=-1
\end{enumerate}

\subsubsection{Region-Based Weighting}

Metrics exhibiting variability across the operational region (i–iv) are aggregated using a weighted average and variance over a grid of foot orientations. To this end, we define two nested rectangular regions: \looseness=-1

\begin{itemize}
    \item a \emph{core region}, which fully encloses the union of all task trajectories and assigns a uniform weight \( w = 1 \);
    \item an \emph{extended region}, defined as the remainder of the operational region used during optimization, where \( w \) smoothly decays to zero via a raised cosine taper.
\end{itemize}

This formulation allows for gradual degradation outside the core region, while still penalizing excessive variance or poor performance in regions more critical to the task set.

Each metric in this category is characterized by both a weighted average $\mu$ and a corresponding variance $\sigma^2$:
\begin{equation*}
    \mu \triangleq \frac{\sum_k w_k\, m_k}{\sum_k w_k}, \quad
    \sigma^2 \triangleq \frac{\sum_k w_k\, (m_k - \mu)^2}{\sum_k w_k},
\end{equation*}
where \( m_k \) denotes the metric value at the \( k \)-th grid point and \( w_k \) is the corresponding weight.

\begin{figure*}[th]
  \centering
  \begin{subfigure}[b]{0.16\textwidth}
    \includegraphics[width=\linewidth]{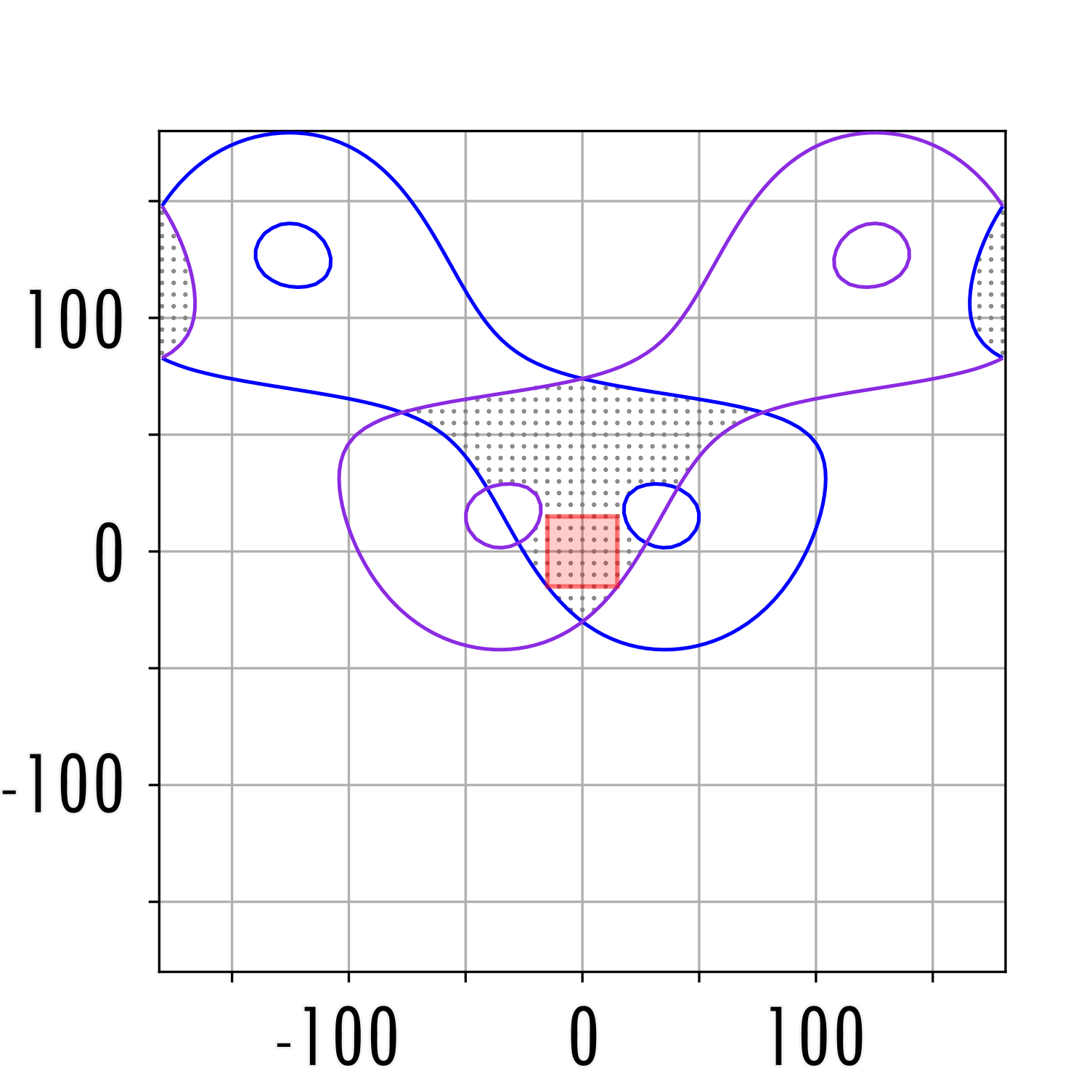}
  \end{subfigure}
  \begin{subfigure}[b]{0.16\textwidth}
    \includegraphics[width=\linewidth]{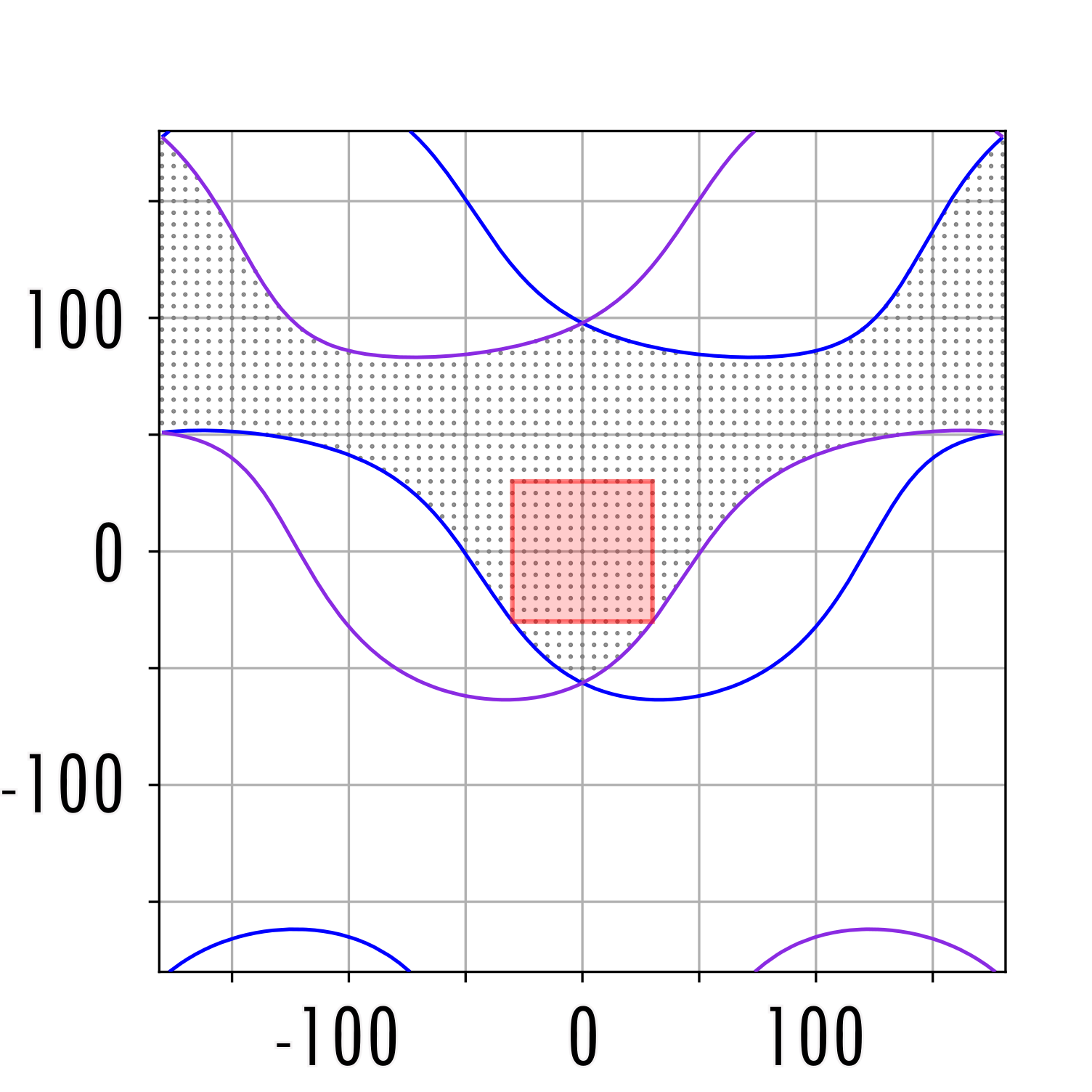}
  \end{subfigure}
    \begin{subfigure}[b]{0.16\textwidth}
    \includegraphics[width=\linewidth]{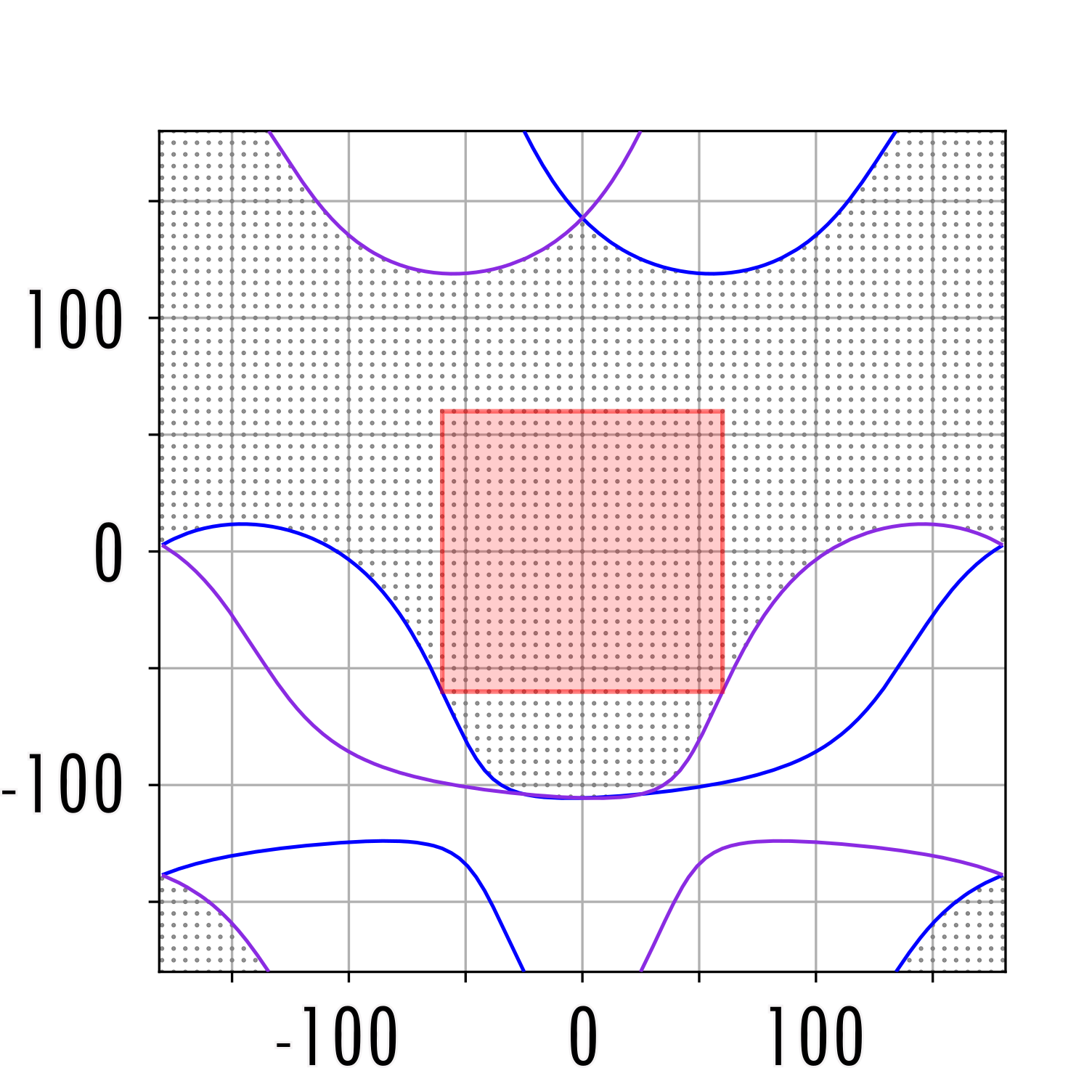}
  \end{subfigure}
    \begin{subfigure}[b]{0.16\textwidth}
    \includegraphics[width=\linewidth]{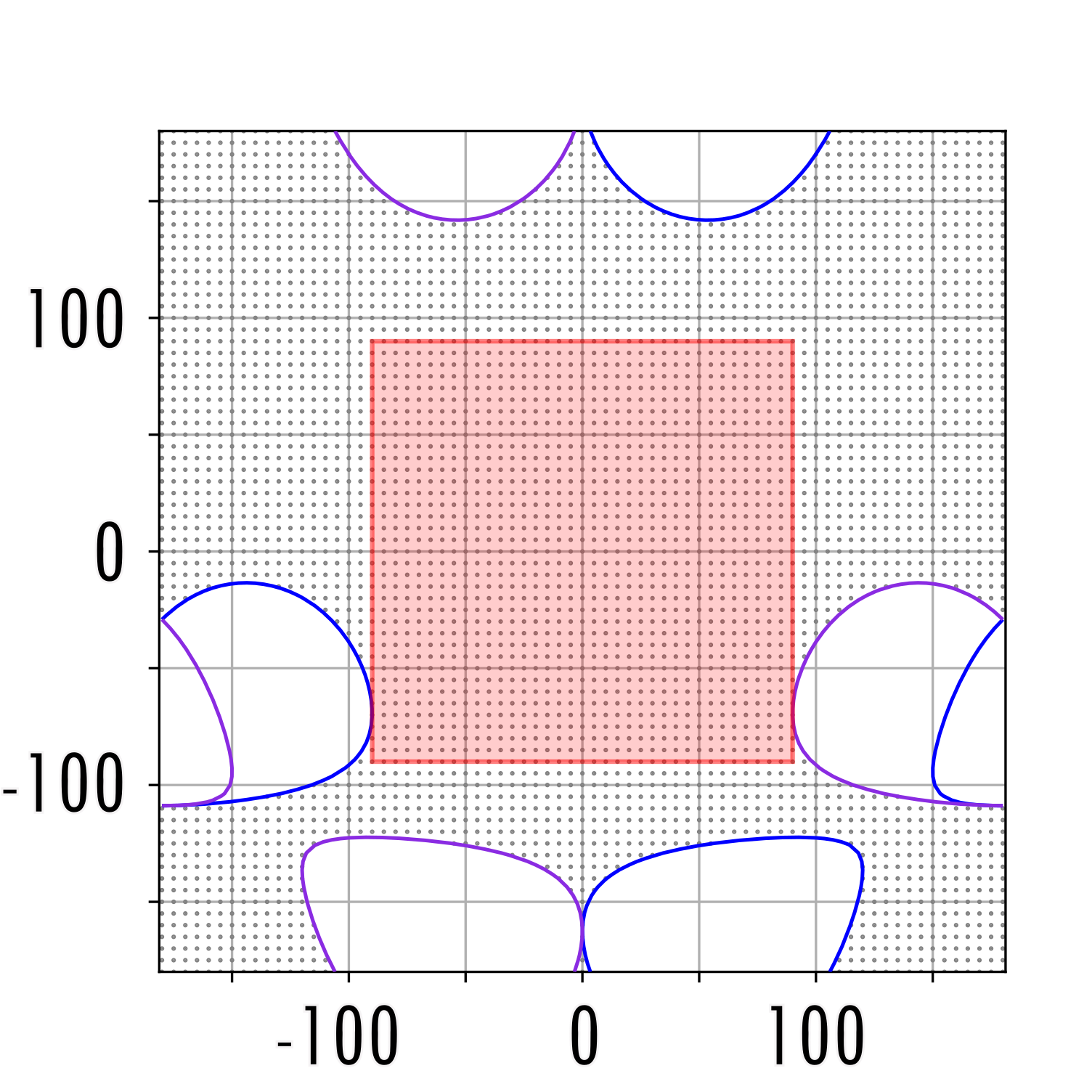}
  \end{subfigure}
    \begin{subfigure}[b]{0.16\textwidth}
    \includegraphics[width=\linewidth]{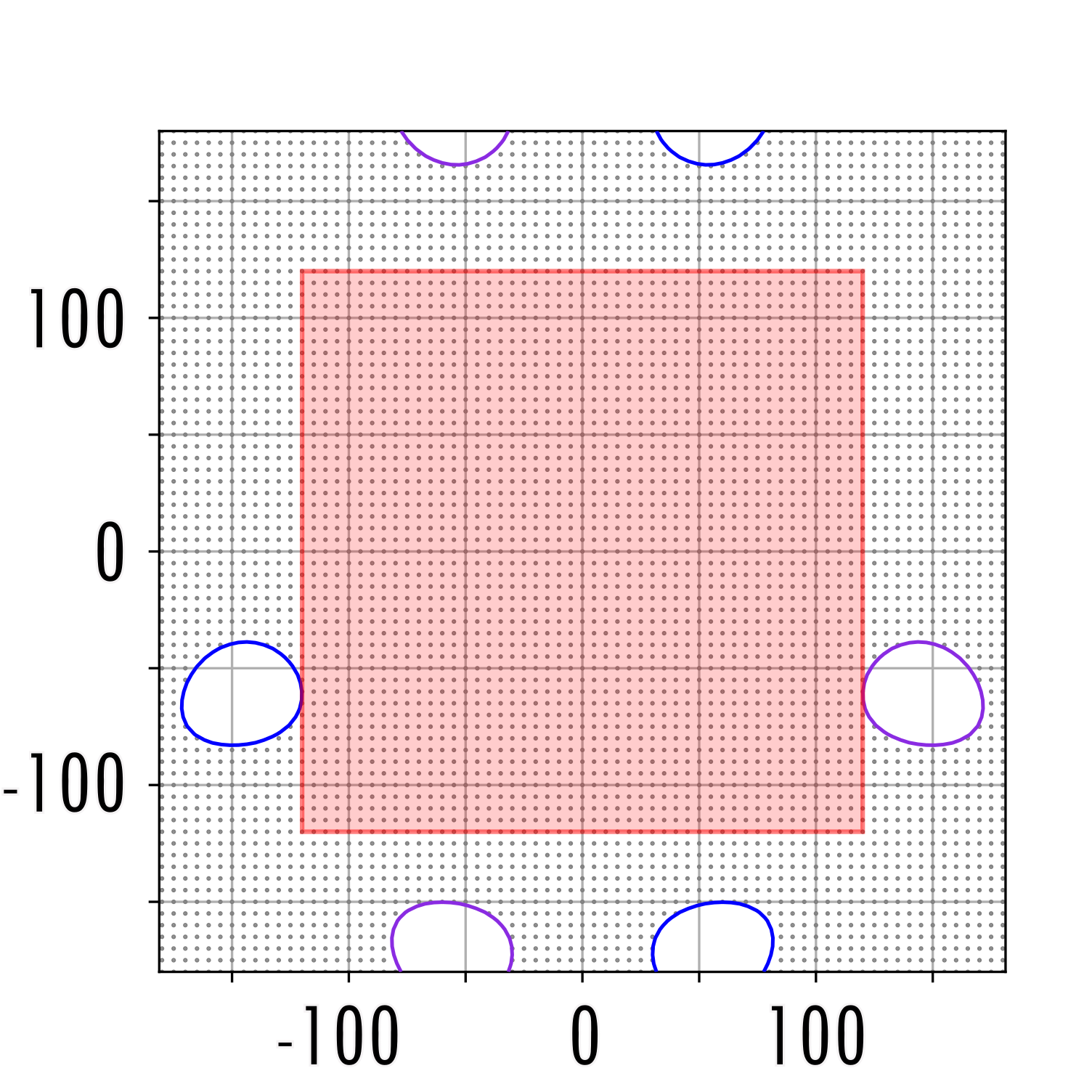}
  \end{subfigure}
    \begin{subfigure}[b]{0.16\textwidth}
    \includegraphics[width=\linewidth]{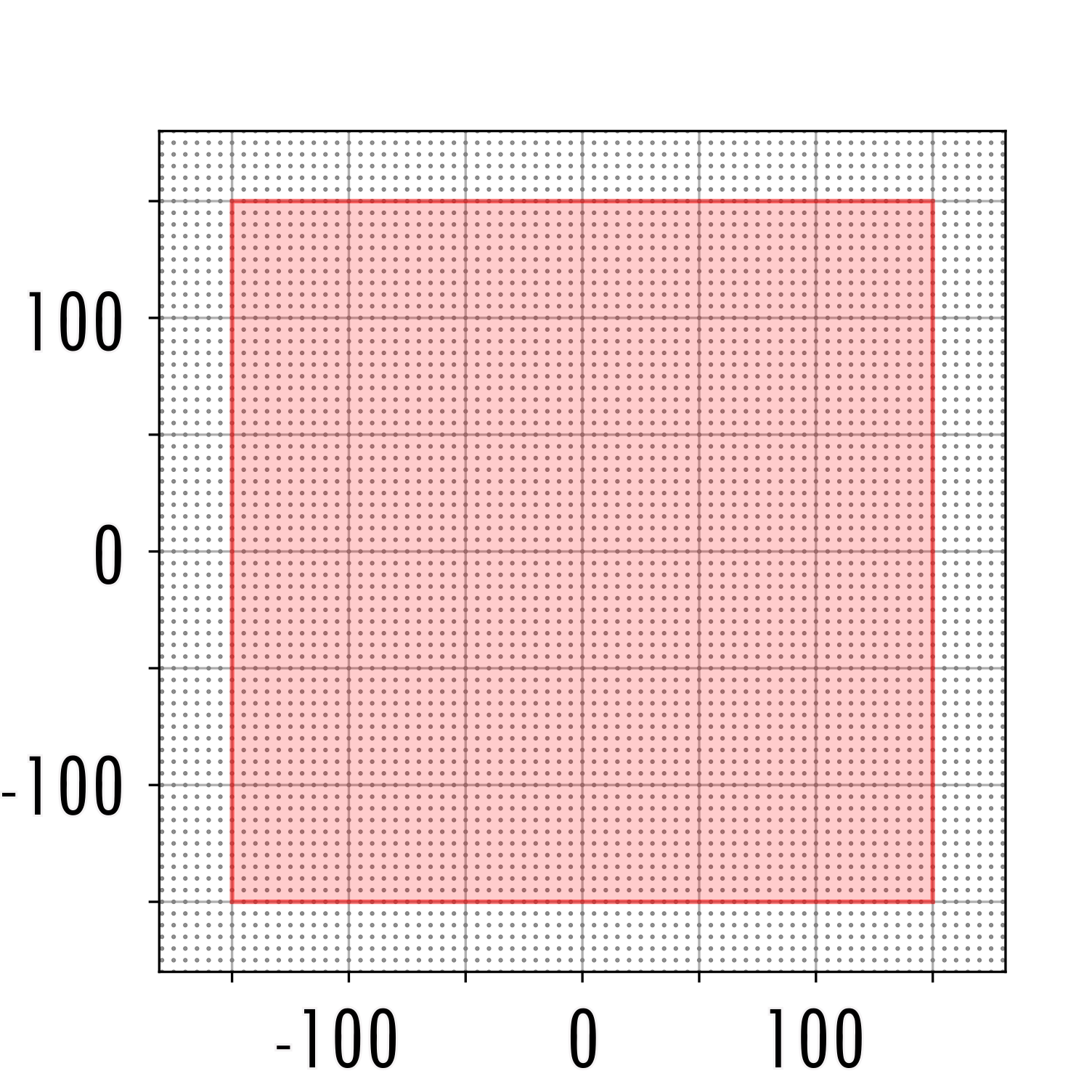}
  \end{subfigure}
  \caption{RSU Reparameterization. The red region is the desired operational region, the gray region is the configuration space of the mechanism (roll on the $x$-axis, pitch on the $y$-axis). The curves represent the configurations where the crank and the rod are aligned (leg 1 in blue, leg 2 in purple). The white region denotes configurations for which the IK problem admits no solution, i.e., the mechanism cannot form a closed kinematic chain. The red region is always contained within the gray region. The parameters used for this analysis are:
    ${}^\mathcal{W}\bm{a}_1 = [-86,\, 40,\, 235]^\top \text{mm}$,
    ${}^\mathcal{W}\bm{a}_2 = [-86,\, -40,\, 235]^\top \text{mm}$, 
    ${}^\mathcal{F}\bm{b}_1 = [-34,\, 36,\, 36]^\top \text{mm}$,
    ${}^\mathcal{F}\bm{b}_2 = [-34,\, -36,\, 36]^\top \text{mm}$,
    $\psi_1 = -90^\circ$,
    $\psi_2 = 90^\circ$,
    $\gamma_1 = \gamma_2 = 0.001$,
    $\delta_1 = \delta_2 = 0.001$.
  }
  \vspace{-1.5em}
  \label{fig:reparameterization}
\end{figure*}
\subsubsection{Cost Function Definition}
Each performance metric is first normalized to the $[0,\, 1]$ range using min-max scaling, computed over the entire population of candidates across all actuators. Let $m_j$ denote the raw value of the $j$-th metric for a given individual, and let $m_{j, \text{min}}$ and $m_{j, \text{max}}$ be the minimum and maximum values of that metric over the full candidate set. The normalized value $\tilde{m}_j$ is then defined as:
\begin{equation*}
    \tilde{m}_j {\triangleq}
    \begin{cases}
        \displaystyle \frac{m_j {-} m_{j,\, \text{min}}}{m_{j,\, \text{max}} {-} m_{j,\, \text{min}}} & \text{if low-metrics are preferable}, \\
        \displaystyle \frac{m_{j,\, \text{max}} {-} m_j}{m_{j,\, \text{max}} {-} m_{j,\, \text{min}}} & \text{if high-metrics are preferable},
    \end{cases}
\end{equation*}
ensuring a common scaling across all mechanism-actuator combinations, making the final scores directly comparable. \looseness=-1

The total cost $\xi$ is computed as a weighted sum:
\begin{equation}
    \label{eq:scala-cost}
    \xi \triangleq \sum\nolimits_{j} \eta_j\, \tilde{m}_j, \quad \text{with} \quad \sum\nolimits_{j} \eta_j = 1.
\end{equation}
The solution with the lowest $\xi$ is deemed most suitable according to the specified performance priorities.

Weights $\eta_j$ can be tuned depending on application-specific preferences (e.g., prioritize backdrivability or compactness).

\section{Results}
\label{sec:results}
We first validate the RSU parameterization, then we test the evaluation framework, and finally, we compare the proposed design with an engineered RSU and a serial design.\looseness=-1

\subsection{Validation of the RSU Reparameterization}
In Fig.~\ref{fig:reparameterization}, a series of tests is presented to validate the correctness of the proposed approach. Given a set of parameters for the mechanism and a desired operational region (which is progressively varied in the figure from $[{-}15^\circ, 15^\circ] {\times} [{-}15^\circ, 15^\circ]$ to $[{-}150^\circ, 150^\circ] {\times} [{-}150^\circ, 150^\circ]$), it is verified that this region remains entirely within the area where IK is solvable. The operational region touches the singularity curves because values approaching zero were used for $\gamma_i$ and $\delta_i$, placing the system in limiting conditions. \looseness=-1

\subsection{Mechanisms Evaluation Framework Use-case}
\begin{figure}[t]
    \centering
    \includegraphics[width=0.75\linewidth]{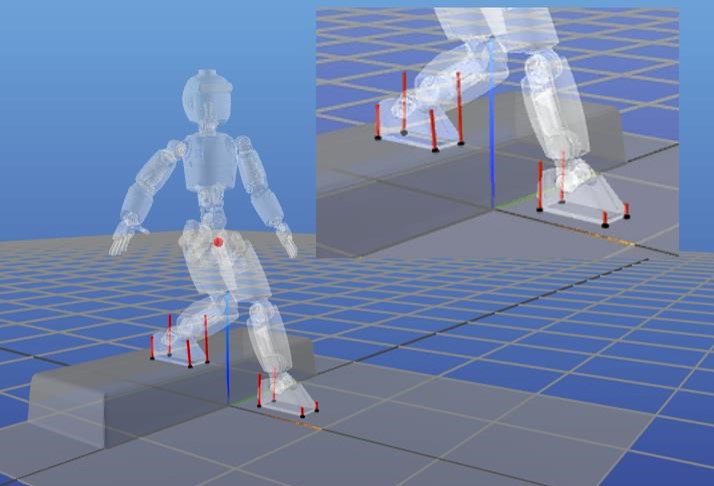}
    \caption{Reference task example. The robot executes a step-down maneuver generated via trajectory optimization.}
    \vspace{-1.2em}
    \label{fig:task}
\end{figure}

\begin{figure}[t]    
    \centering
    \includegraphics[width=0.95\linewidth]{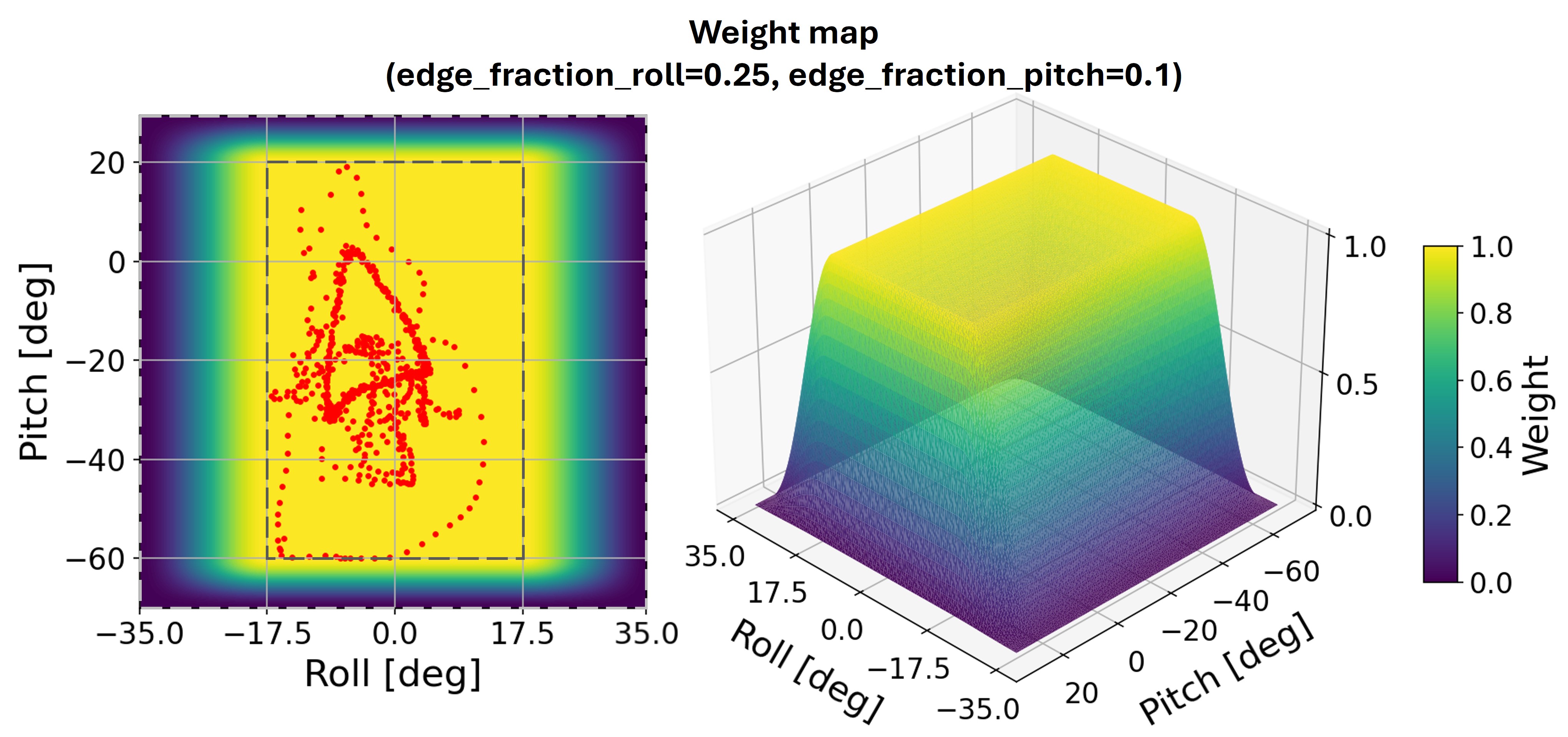}
    \caption{Raised cosine tapered weight map. Red dots mark the foot configurations observed during the simulated and experimental tasks.}
    \label{fig:weight_map}
    \vspace{-1.5em}
\end{figure}

The design framework presented in Section~\ref{sec:kinematic_analysis} is implemented in Grasshopper and Python. The multi-objective optimization problem is solved with the NSGA-II algorithm~\cite{deb2002fast} using Tunny plugin\cite{natsume2024tunny}.
The framework is applied to evaluate the upgrade of the serial ankle of a humanoid robot into parallel architectures capable of executing a broader set of motion tasks. The desired tasks are defined by a mix of experimental and simulated data, including: (i) a dynamic walking trajectory obtained from experiments; (ii) simulated walking on a ramp with 20\% inclination; and (iii) simulated walking up and down a 20~cm step (depicted in Fig.\ref{fig:task}). The union of the ankle roll and pitch trajectories across these tasks led to the definition of a required \emph{operational region} for the new ankle, set to $[-35^\circ,\, 35^\circ]$ in roll and $[-70^\circ,\, 30^\circ]$ in pitch. A more compact \emph{core region}, defined as $[-17.5^\circ,\, 17.5^\circ]$ in roll and $[-60^\circ,\, 20^\circ]$ in pitch, was used to guide the region-based weighting introduced in Sec.~\ref{subsec:cost_function} -- depicted in Fig.~\ref{fig:weight_map}.\looseness=-1

In our study, we chose to design the mechanism using commercial actuators. To estimate the minimum mechanical power required to perform the target tasks, we conducted a preliminary exploration of the design space through multi-objective optimizations.
This process enabled us to assess the power demands associated with different design configurations and informed the selection of suitable commercial actuators, specifically the linear actuator Wittenstein AL32 (ball screw, 2mm lead, 160mm stroke) for the SPU, and the rotary actuators Maxon HEJ 70-48-50, MyActuator RMD-X6-P20-60, and Synapticon ACTILINK JD 10 for the RSU. The rotary actuators feature low-reduction planetary gearboxes (ratios 9:1--19:1), selected to promote high backdrivability.

\begin{figure}[t]
    \centering   
    \includegraphics[width=0.85\linewidth]{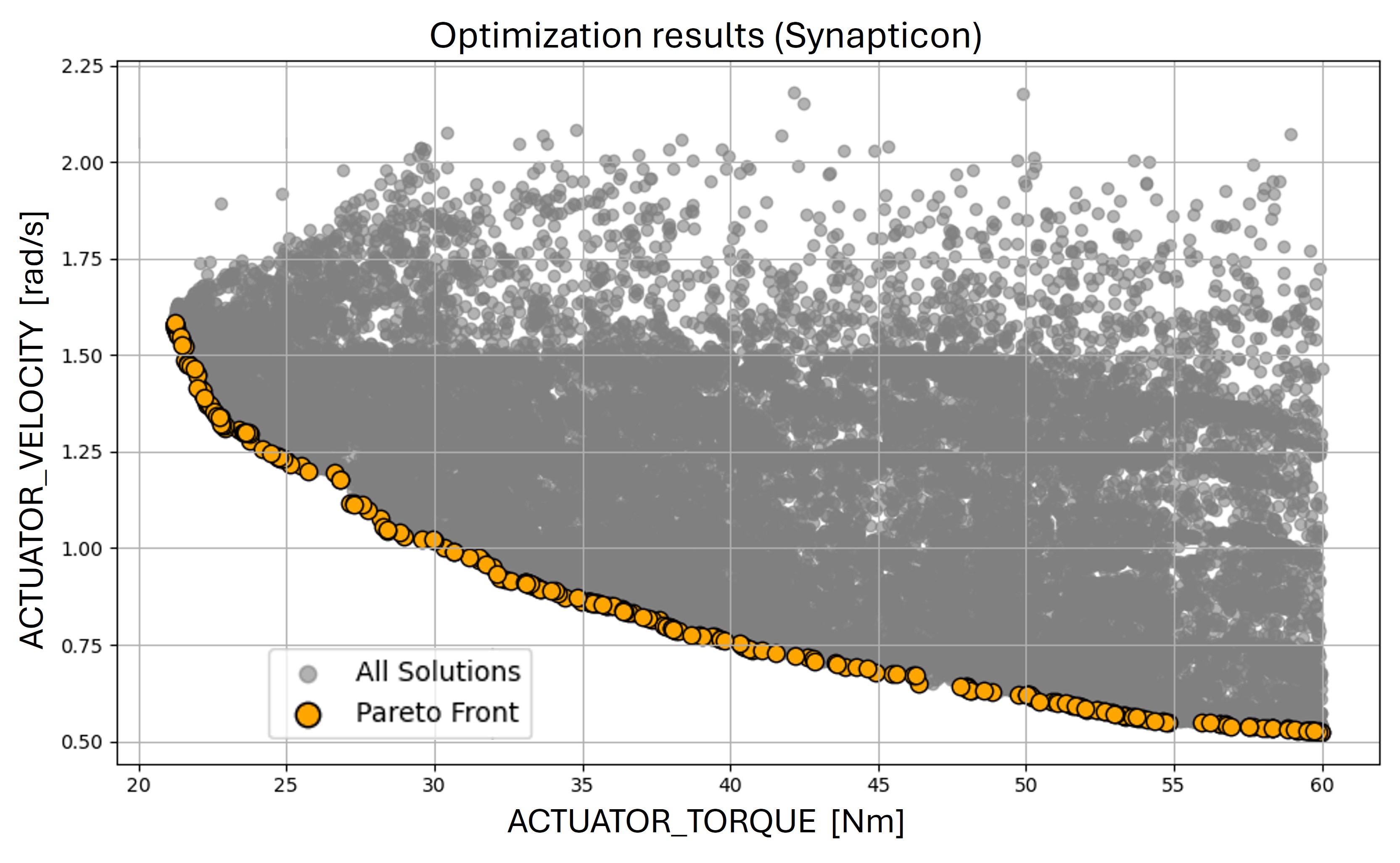}
    \caption{Output of the multi-objective optimization problem with Synapticon actuators and RSU mechanism.}
    \label{fig:pareto_front_synapticon}
    \vspace{-1.2em}
\end{figure}

\begin{figure}[t]
    \centering   
    \includegraphics[width=0.8\linewidth]{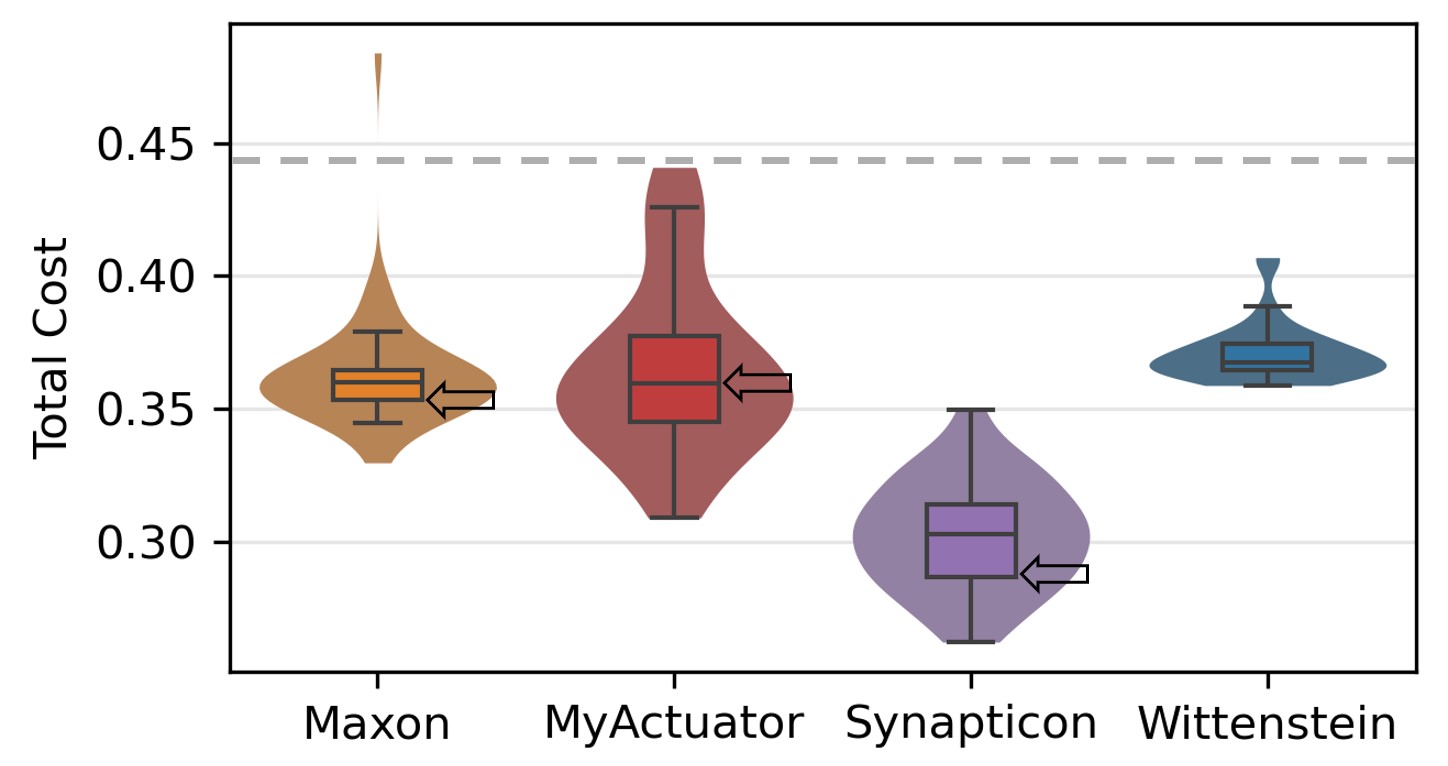}
    \caption{Cost distribution $\xi$ of the Pareto fronts for each actuator. Arrows indicate the performance of the engineered RSU mechanism, while the dashed line represents the performance of the original serial architecture.\looseness=-1}
    \label{fig:total_cost}
    \vspace{-1.5em}
\end{figure}

For each actuator, a refined design space $\Pi$ is defined by introducing geometric constraints to ensure collision-free integration. The multi-objective optimization problem described in~\eqref{eq:optimisation-problem} is solved, resulting in a Pareto front -- see Fig.\ref{fig:pareto_front_synapticon}.
For each solution on the Pareto front, a scalar cost $\xi$ is computed according to \eqref{eq:scala-cost}, using uniformly distributed weights $\eta_j$ across the seven performance metrics. The resulting cost values are summarized in Fig.~\ref{fig:total_cost}, grouped by actuator family. The plot reveals a consistent performance advantage for solutions involving the Synapticon actuator, with lower average cost and tighter variance.\looseness=-1

\begin{figure}[t]
    \vspace{0.5em}
    \centering   
    \includegraphics[width=0.94\linewidth]{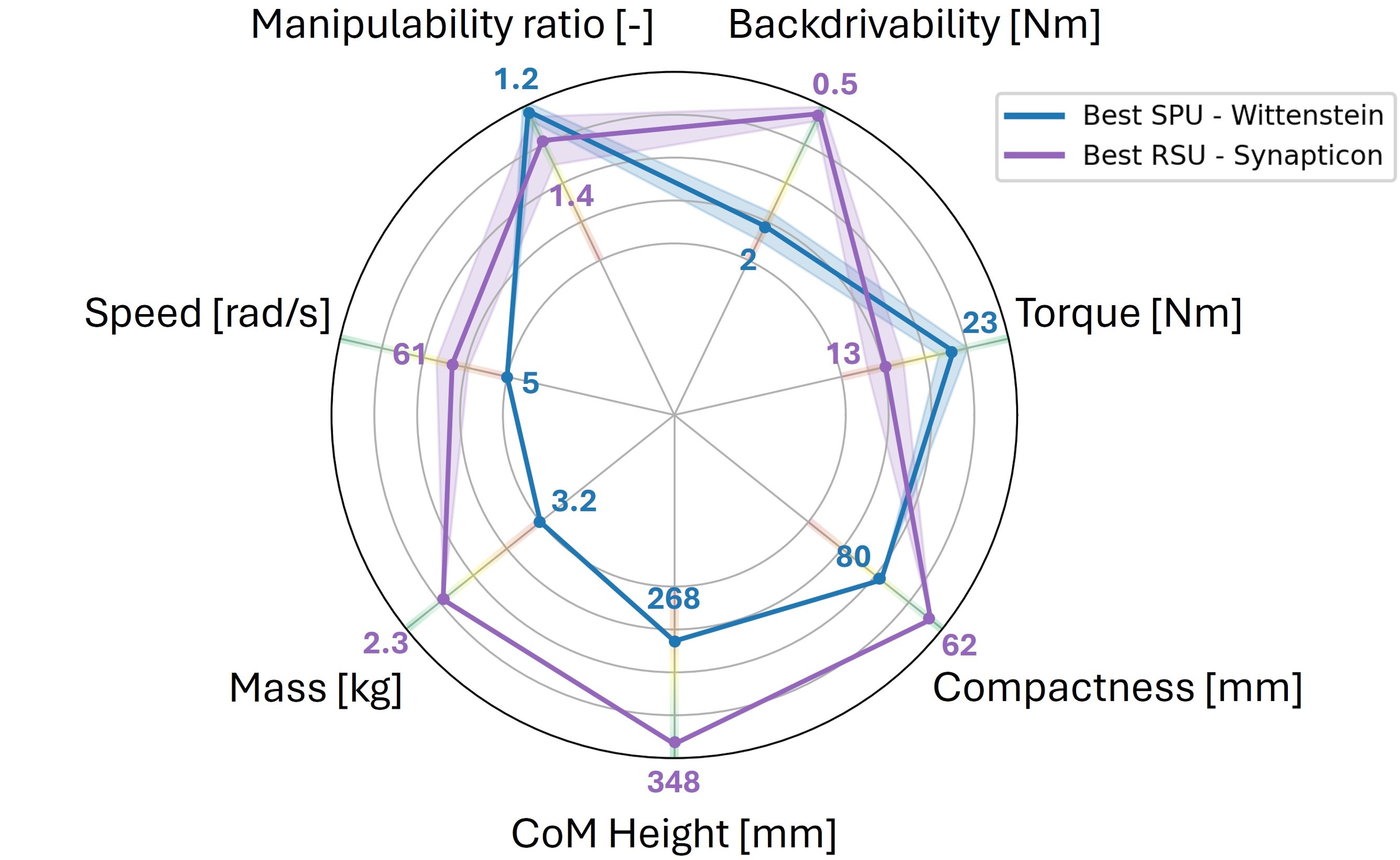}
    \vspace{0.5em}
    \caption{Comparison of the best SPU and RSU candidates. The solid line is the average value $\mu$, while the shaded area represents the standard deviation $\sigma$ over the evaluated range of motion. Values along the seven axes are individually min–max normalized for readability.\looseness=-1}
    \label{fig:spu_vs_rsu}
    \vspace{-1.5em}
\end{figure}

To gain further insight into the structure of the cost metric, we selected the best SPU and RSU designs and visualized their performance breakdown across all seven metrics. As shown in Fig.~\ref{fig:spu_vs_rsu}, the RSU candidate outperforms the SPU counterpart across most metrics. The radial width of the bands instead reveals that the SPU design exhibits lower performance variability across the operational region. Despite this trade-off, the RSU remains the more favorable solution due to its overall superior capabilities.\looseness=-1


\subsection{Validation of the scalar cost function approach}
We validated our approach against a conventionally engineered RSU (Fig.~\ref{fig:SPU_RSU_render}), designed via iterative CAD refinement to ensure coverage of the operational region, avoid collisions, and meet joint torque/speed requirements. While carefully constructed, this process lacks global optimality guarantees. 

Given the achieved geometry, we paired it with each rotary actuator used in the RSU study, yielding three additional candidates. Their total costs $\xi$ are shown as arrows in Fig.~\ref{fig:total_cost}. In all cases, the optimized populations include solutions with strictly lower cost, confirming the advantage of our automated pipeline. \looseness=-1

\subsection{Comparison with the original serial architecture}
To assess the improvement over the studied existing ankle, we computed the total cost $\xi$ of its current serial architecture, shown as a dashed line in Fig.~\ref{fig:total_cost}. This not only validates nearly all optimized designs, but also demonstrates that our method applies to serial mechanisms as well. In this case, the region-based weighting from Section~\ref{subsec:cost_function} is unnecessary, as serial architectures exhibit configuration-independent performance.

\section{Conclusions}
\label{sec:conclusions}
This paper presented a unified framework for the design and evaluation of parallel ankle mechanisms in humanoid robots, enabling informed, quantitative comparisons across architectures and actuators. The pipeline integrates closed-form inverse kinematics, a novel RSU reparameterization that guarantees workspace feasibility, task-informed multi-objective optimization, and a scalar cost function for structured evaluation. \looseness=-1

We applied the framework to two widely adopted architectures—SPU and RSU—modeled from both topological and geometrical perspectives. In a realistic redesign scenario, the optimized RSU reduced the scalar cost by 41\% over the original serial design and by 14\% over a conventionally engineered RSU, demonstrating the effectiveness of the proposed approach.

Future work will address collision checking over the full operational region and embedding the cost function directly into the optimization loop for performance-guided search.

\bibliographystyle{IEEEtran}
\bibliography{IEEEabrv, bibliography}

\begin{thebibliography}{10}
\providecommand{\url}[1]{#1}
\csname url@samestyle\endcsname
\providecommand{\newblock}{\relax}
\providecommand{\bibinfo}[2]{#2}
\providecommand{\BIBentrySTDinterwordspacing}{\spaceskip=0pt\relax}
\providecommand{\BIBentryALTinterwordstretchfactor}{4}
\providecommand{\BIBentryALTinterwordspacing}{\spaceskip=\fontdimen2\font plus
\BIBentryALTinterwordstretchfactor\fontdimen3\font minus \fontdimen4\font\relax}
\providecommand{\BIBforeignlanguage}[2]{{%
\expandafter\ifx\csname l@#1\endcsname\relax
\typeout{** WARNING: IEEEtran.bst: No hyphenation pattern has been}%
\typeout{** loaded for the language `#1'. Using the pattern for}%
\typeout{** the default language instead.}%
\else
\language=\csname l@#1\endcsname
\fi
#2}}
\providecommand{\BIBdecl}{\relax}
\BIBdecl

\bibitem{kim2024hyperleg}
D.-Y. Kim, S.-H. Yun, J.-K. Lee, J.~Yoon, D.~Nam, C.-Y. Maeng, and Y.-J. Kim, ``Hyperleg: Biomechanics-inspired high-dof leg and toe mechanism for highly dynamic motions,'' in \emph{2024 IEEE International Conference on Robotics and Automation (ICRA)}.\hskip 1em plus 0.5em minus 0.4em\relax IEEE, 2024, pp. 2456--2462.

\bibitem{Choi2006}
Y.~Choi, D.~Kim, and B.-J. You, ``On the walking control for humanoid robot based on the kinematic resolution of com jacobian with embedded motion,'' in \emph{Proceedings 2006 IEEE International Conference on Robotics and Automation, 2006. ICRA 2006.}, 2006, pp. 2655--2660.

\bibitem{gim2022implementation}
K.~G. Gim and J.~Kim, ``Implementation of a large-scale biped robot using serial-parallel hybrid leg mechanism,'' in \emph{2022 19th International Conference on Ubiquitous Robots (UR)}.\hskip 1em plus 0.5em minus 0.4em\relax IEEE, 2022, pp. 115--121.

\bibitem{zhou2018comprehensive}
C.~Zhou and N.~Tsagarakis, ``On the comprehensive kinematics analysis of a humanoid parallel ankle mechanism,'' \emph{Journal of Mechanisms and Robotics}, vol.~10, no.~5, p. 051015, 2018.

\bibitem{alfayad2009new}
S.~Alfayad, F.~B. Ouezdou, and F.~Namoun, ``New three dof ankle mechanism for humanoid robotic application: Modeling, design and realization,'' in \emph{2009 IEEE/RSJ International Conference on Intelligent Robots and Systems}.\hskip 1em plus 0.5em minus 0.4em\relax IEEE, 2009, pp. 4969--4976.

\bibitem{lohmeier2009humanoid}
S.~Lohmeier, T.~Buschmann, and H.~Ulbrich, ``Humanoid robot lola,'' in \emph{2009 IEEE International Conference on Robotics and Automation}.\hskip 1em plus 0.5em minus 0.4em\relax IEEE, 2009, pp. 775--780.

\bibitem{fuge2023design}
A.~J. Fuge, C.~W. Herron, B.~C. Beiter, B.~Kalita, and A.~Leonessa, ``Design, development, and analysis of the lower body of next-generation 3d-printed humanoid research platform: Pandora,'' \emph{Robotica}, vol.~41, no.~7, pp. 2177--2206, 2023.

\bibitem{roig2022hardware}
A.~Roig, S.~K. Kothakota, N.~Miguel, P.~Fernbach, E.~M. Hoffman, and L.~Marchionni, ``On the hardware design and control architecture of the humanoid robot kangaroo,'' in \emph{6th workshop on legged robots during the international conference on robotics and automation (ICRA 2022)}, 2022.

\bibitem{perera2024staccatoe}
N.~Perera, S.~Yu, D.~Marew, M.~Tang, K.~Suzuki, A.~McCormack, S.~Zhu, Y.-J. Kim, and D.~Kim, ``Staccatoe: A single-leg robot that mimics the human leg and toe,'' in \emph{2024 IEEE/RSJ International Conference on Intelligent Robots and Systems (IROS)}.\hskip 1em plus 0.5em minus 0.4em\relax IEEE, 2024, pp. 9058--9065.

\bibitem{shin2022design}
Y.-H. Shin, S.~Hong, S.~Woo, J.~Choe, H.~Son, G.~Kim, J.-H. Kim, K.~Lee, J.~Hwangbo, and H.-W. Park, ``Design of kaist hound, a quadruped robot platform for fast and efficient locomotion with mixed-integer nonlinear optimization of a gear train,'' in \emph{2022 International Conference on Robotics and Automation (ICRA)}.\hskip 1em plus 0.5em minus 0.4em\relax IEEE, 2022, pp. 6614--6620.

\bibitem{WU20141377}
G.~Wu, S.~Caro, S.~Bai, and J.~Kepler, ``Dynamic modeling and design optimization of a 3-dof spherical parallel manipulator,'' \emph{Robotics and Autonomous Systems}, vol.~62, no.~10, pp. 1377--1386, 2014.

\bibitem{YANG2022104725}
C.~Yang, W.~Ye, and Q.~Li, ``Review of the performance optimization of parallel manipulators,'' \emph{Mechanism and Machine Theory}, vol. 170, p. 104725, 2022.

\bibitem{semini2016design}
C.~Semini, V.~Barasuol, J.~Goldsmith, M.~Frigerio, M.~Focchi, Y.~Gao, and D.~G. Caldwell, ``Design of the hydraulically actuated, torque-controlled quadruped robot hyq2max,'' \emph{IEEE/Asme Transactions on Mechatronics}, vol.~22, no.~2, pp. 635--646, 2016.

\bibitem{ivolga2023computational}
D.~V. Ivolga, I.~I. Borisov, K.~V. Nasonov, and S.~A. Kolyubin, ``Computational design of closed-chain linkages: Respawn algorithm for generative design,'' in \emph{2023 IEEE/RSJ International Conference on Intelligent Robots and Systems (IROS)}.\hskip 1em plus 0.5em minus 0.4em\relax IEEE, 2023, pp. 481--486.

\bibitem{pierrot2009optimal}
F.~Pierrot, V.~Nabat, O.~Company, S.~Krut, and P.~Poignet, ``Optimal design of a 4-dof parallel manipulator: From academia to industry,'' \emph{IEEE Transactions on Robotics}, vol.~25, no.~2, pp. 213--224, 2009.

\bibitem{hashimoto2020mechanics}
K.~Hashimoto, ``Mechanics of humanoid robot,'' \emph{Advanced Robotics}, vol.~34, no. 21-22, pp. 1390--1397, 2020.

\bibitem{ficht2021bipedal}
G.~Ficht and S.~Behnke, ``Bipedal humanoid hardware design: A technology review,'' \emph{Current Robotics Reports}, vol.~2, pp. 201--210, 2021.

\bibitem{huang2012theory}
Z.~Huang, Q.~Li, and H.~Ding, \emph{Theory of parallel mechanisms}.\hskip 1em plus 0.5em minus 0.4em\relax Springer Science \& Business Media, 2012, vol.~6.

\bibitem{yamane2019closed}
K.~Yamane, \emph{Closed-Loop Dynamics}.\hskip 1em plus 0.5em minus 0.4em\relax Berlin, Heidelberg: Springer Berlin Heidelberg, 2019, pp. 1--6.

\bibitem{liu2014parallel}
X.-J. Liu and J.~Wang, ``Parallel kinematics,'' \emph{Springer Tracts in Mechanical Engineering}, 2014.

\bibitem{lynch2017modern}
K.~M. Lynch and F.~C. Park, \emph{Modern robotics}.\hskip 1em plus 0.5em minus 0.4em\relax Cambridge University Press, 2017.

\bibitem{gosselin1990singularity}
C.~Gosselin, J.~Angeles \emph{et~al.}, ``Singularity analysis of closed-loop kinematic chains.'' \emph{IEEE transactions on robotics and automation}, vol.~6, no.~3, pp. 281--290, 1990.

\bibitem{yoshikawa1985manipulability}
T.~Yoshikawa, ``Manipulability of robotic mechanisms,'' \emph{The international journal of Robotics Research}, vol.~4, no.~2, pp. 3--9, 1985.

\bibitem{gogu2005chebychev}
G.~Gogu, ``Chebychev--gr{\"u}bler--kutzbach's criterion for mobility calculation of multi-loop mechanisms revisited via theory of linear transformations,'' \emph{European Journal of Mechanics-A/Solids}, vol.~24, no.~3, pp. 427--441, 2005.

\bibitem{deb2002fast}
K.~Deb, A.~Pratap, S.~Agarwal, and T.~Meyarivan, ``A fast and elitist multiobjective genetic algorithm: Nsga-ii,'' \emph{IEEE Transactions on Evolutionary Computation}, vol.~6, no.~2, pp. 182--197, 2002.

\bibitem{natsume2024tunny}
\BIBentryALTinterwordspacing
H.~Natsume, ``{Tunny, The Grasshopper optimization tool},'' Jun. 2024. [Online]. Available: \url{https://github.com/hrntsm/Tunny}
\BIBentrySTDinterwordspacing

\end{thebibliography}
\end{document}